\documentclass[11pt]{article}
\usepackage{eamt24}
\usepackage{times}
\usepackage{url}
\usepackage{diagbox}
\usepackage{latexsym}
\usepackage[T1]{fontenc}
\usepackage[small,bf]{caption} 
\usepackage{amssymb}
\usepackage{multirow}
\usepackage[dvipsnames]{xcolor}
\usepackage{todonotes}
\title{FAME-MT Dataset: Formality Awareness Made Easy for Machine Translation Purposes}

\author{Dawid Wiśniewski$\normalfont{\textsuperscript{1,2}}$, Zofia Rostek$\normalfont{\textsuperscript{1}}$, Artur Nowakowski$\normalfont{\textsuperscript{1,3}}$ \\
$\textsuperscript{1}$ Laniqo, Poznań, Poland \\
$\textsuperscript{2}$ Faculty of Computing and Telecommunications, Poznań University of Technology, Poland \\
$\textsuperscript{3}$ Faculty of Mathematics and Computer Science, Adam Mickiewicz University, Poznań, Poland \\
{\tt {name}.{surname}@laniqo.com}}
\date{}

\begin{document}
\maketitle
\begin{abstract}
People use language for various purposes. Apart from sharing information, individuals may use it to express emotions or to show respect for another person.
In this paper, we focus on the formality level of machine-generated translations and present \textbf{FAME-MT} -- a dataset consisting of 11.2 million translations between 15 European source languages and 8 European target languages classified to formal and informal classes according to target sentence formality. This dataset can be used to fine-tune machine translation models to ensure a given formality level for each European target language considered. We describe the dataset creation procedure, the analysis of the dataset's quality showing that \textbf{FAME-MT} is a reliable source of language register information, and we present a publicly available proof-of-concept machine translation model that uses the dataset to steer the formality level of the translation. Currently, it is the largest dataset of formality annotations, with examples expressed in 112 European language pairs. The dataset is published online\footnote{\url{https://github.com/laniqo-public/fame-mt/}}.
\end{abstract}






\section{Introduction}
\paragraph{Motivation} 
Different situations require using different language depending on whether it is a formal meeting or a casual talk with friends. Frequently, when speaking to an older or distinguished person, e.g., an owner of a company or a university professor, we use appropriate language forms to show respect. However, machine translation models may struggle with choosing an appropriate language form due to lack of context -- often, we only have one sentence to translate, there are cultural differences between source and target language speakers, and models may be trained on parallel corpora, which do not focus on formality aspect of language enough.

For this reason, it is important to find ways of enforcing machine translation models to use a required formality level in translated sequences. While there are methods that can be used to achieve this goal, the existing datasets are scarce, focusing either on formality classification for a few selected languages (e.g., German or English) or providing formality-annotated examples of translations between pairs of languages, which are limited in size and include a narrow selection of target languages with English as the source language. What we found missing is a large-scale dataset providing formality-annotated examples of translations for a much broader set of languages, enabling easy fine-tuning of pre-trained machine translation models not only for directions including English but also other European languages.

\paragraph{Contribution} In this paper, we present \textbf{FAME-MT}, the biggest to date dataset of translation pairs annotated with formality level. This dataset introduces 100,000 annotated translation examples for each of the 112 European language pairs considered. With 8 target languages (English, German, French, Italian, Dutch, Polish, Portuguese, and Spanish) classified according to formality and 15 European languages considered as the source languages, the dataset introduces an 18-fold increase of the European language-pair coverage over the most diverse dataset \textbf{CoCoA-MT}~\cite{DBLP:conf/naacl/NadejdeCHNFD22} and over 100-fold increase of the dataset size considering the biggest formality-annotated datasets available~\cite{DBLP:conf/naacl/RaoT18}. 

Providing translation examples for a wide selection of source and target languages, \textbf{FAME-MT} allows for simple fine-tuning of machine translation models for the most popular European language pairs. 

In this paper, we discuss the dataset creation process and analyze the data quality using various metrics to prove the usefulness of \textbf{FAME-MT}. We show how proof-of-concept models for formality-aware machine translation can be trained using Marian~\cite{mariannmt}, provide examples of outputs of those models, and publish models online along with the dataset.

\paragraph{Research questions}
We formulated the following research questions to be answered in this paper: 
\begin{itemize}
    \item \textbf{RQ1}: Is it possible to create a good quality large-scale dataset for formality-aware machine translation automatically based on available resources?
    \item \textbf{RQ2}: Is 100,000 translation examples for a given language pair enough to fine-tune a pre-trained machine translation model to become formality-aware?
    \item \textbf{RQ3}: Considering translation pairs coming from \textbf{MTData}~\cite{gowda-etal-2021-many}, are formal sentences always translated into formal ones, and informal sentences translated into informal ones?
\end{itemize}

\section{Related work} 

The idea of identifying formality level in texts is a widely analyzed area of (socio)linguistics~\cite{biber2019register}. In linguistics, there are 5 main language registers defined that can be used in particular situations. These are:
frozen, formal, consultative, casual (informal), and intimate registers. 
Out of them, the formal and informal ones are most often analyzed in the context of machine translation.

\paragraph{Existing formality datasets} There are several datasets proposed that help incorporate formality awareness in various NLP tasks. The biggest one, Grammarly’s Yahoo Answers Formality Corpus (or \textbf{GYAFC} corpus for short~\cite{DBLP:conf/naacl/RaoT18}), introduces 110,000 pairs of formal and informal sentences in English. The dataset is based on Yahoo Answers L6 corpus~\footnote{\url{https://webscope.sandbox.yahoo.com/catalog.php?datatype=l}} and is proposed to be used for style transfer. The authors proposed an LSTM-based baseline showing that style transfer can be achieved using the machine translation approach and propose metrics for automatic evaluation of style-transfer models (e.g., fluency, meaning preservation). 

Another dataset, consisting of 6,574 English sentences annotated with formality level, is proposed by~\newcite{pavlick-tetreault-2016-empirical}. This work is aimed at the analysis of the formality level in English, considering humans’ perceptions of formality in four different genres. 

There are also datasets that do not focus on English, for example, a dataset consisting of 3,000 German sentences annotated by human experts on a continuous scale using comparative judgments~\cite{DBLP:conf/eacl/EderKW23}. Each annotator was presented with several examples, and their goal was to rank the sentences according to formality level. This dataset covers a set of 12 diverse data sources (e.g., Twitter, Reddit, Wikipedia, or Springer Open Science articles).

Apart from the datasets that focus on a single language, there are several datasets with multilingual data. One of them is \textbf{XFORMAL}~\cite{DBLP:conf/naacl/BriakouLZT21}. Similarly to \textbf{GYAFC}, the goal of \textbf{XFORMAL} is to provide a benchmark for style transfer by introducing pairs of informal sentences and their formal counterparts. The dataset provides examples in three languages: French, Italian, and Brazilian-Portuguese, and introduces 1,000 human-annotated examples for each of these languages. However, access to both \textbf{GYAFC} and \textbf{XFORMAL} is limited as it requires access approval as described on the website~\footnote{\url{https://github.com/Elbria/xformal-FoST/}}.

A publicly available dataset called \textbf{CoCoA-MT}~\cite{DBLP:conf/naacl/NadejdeCHNFD22} provides a set of triples consisting of English sentences translated into formal and informal versions for each of the supported languages, namely, German, Spanish, Italian, French, Dutch, Portuguese, Japanese, and Hindi. The dataset provides around 1,000 examples divided into train and test sets for each target language. What distinguishes \textbf{CoCoA-MT} is the presence of phrase annotations that tell which phrases make a given example formal or informal. Even though \textbf{CoCoA-MT} provides translations between language pairs, they are always translations from English to one of the supported languages.

In general, although there are datasets focusing on formality, the large ones (\textbf{GYAFC}) cover only English or German, while those including other languages are relatively small (\textbf{XFORMAL}) and provide translations only from English to a given language (\textbf{CoCoA-MT}). No large-scale multi-language dataset exists up to date.

\paragraph{Formality control methods}
The tasks of style transfer and machine translation can be controlled using various methods. Here, we discuss the most relevant ones in the context of formality control.

The most straightforward method of controlling formality injects a special token in a source sequence, which tells us what level of formality should be achieved on the target side. This method was utilized in various approaches, e.g.: 
    controlling honorifics in English to Japanese translation~\cite{DBLP:conf/aclwat/FeelyHG19} using one of \texttt{\{informal, polite, formal\}} tags injected to the input of the Transformer model, 
    controlling formality of French to English translations using one of predefined formality levels~\cite{DBLP:conf/emnlp/NiuMC17} \texttt{\{low, neutral, high\}},
    controlling formality level by injecting a special token representing one of \texttt{\{informal, formal\}} classes to control output formality presented as part of \textbf{CoCoA-MT} evaluation~\cite{DBLP:conf/naacl/NadejdeCHNFD22},
    controlling politeness of the output text using predefined Latin tokens: \texttt{\{vos, tu\}}~\cite{DBLP:conf/naacl/SennrichHB16},
    or controlling formality by attaching a special token to both input and output sequences for better control of the output~\cite{DBLP:conf/aaai/NiuC20}.

Alternatively, one can add a special embedding vector for each token~\cite{DBLP:conf/emnlp/SchioppaVSF21} to represent the desired formality level, craft prompts~\cite{DBLP:journals/corr/abs-2202-11822} for multilingual T5~\cite{DBLP:conf/naacl/XueCRKASBR21} models, or use Bayesian factorization for constrained output generation~\cite{DBLP:conf/naacl/YangK21}.

\section{Dataset}

The \textbf{FAME-MT} dataset creation process was divided into three steps.

\paragraph{Step 1: Input data selection}
The aim of the \textbf{FAME-MT} project was to provide translations between pairs of languages that are annotated with formality information. For this purpose, the first step was to identify parallel corpora for languages of interest.  We considered eight target languages, i.e., languages into which we want to translate input statements. These are: \textit{English (EN), German (DE), French (FR), Italian (IT), Dutch (NL), Polish (PL), Portuguese (PT), and Spanish (ES)}.  For each target language, our goal was to support a wide selection of source languages, i.e., those from which we translate into the target language.

We selected 15 source languages: \textit{Czech (CS), Danish (DA), German (DE), English (EN), Spanish (ES), French (FR), Italian (IT), Dutch (NL), Norwegian} -- including Norwegian Bokmål \textit{(NO + NB), Polish (PL), Portuguese (PT), Russian (RU), Slovak (SK), Swedish (SV)}, and \textit{Ukrainian (UK)}. This selection of source and target languages resulted in 112 potential language pairs, for each of which we needed a parallel corpus of translations between the given language pair. To collect such a dataset, we used \textbf{MTData} tool that provides an access to machine translation dataset collections, e.g., OPUS~\cite{tiedemann-2016-opus}, gives an access to popular datasets such as Europarl~\cite{koehn-2005-europarl}, or Paracrawl~\cite{banon-etal-2020-paracrawl}, and covers every language pair considered. For each language pair, we acquired a corpus of translation examples from \textbf{MTData} and applied the following postprocessing to increase the quality of the collected data: we rejected documents with more than 15\% of characters being digits, having less than 5 characters in any sentence, having any token longer than 28 characters, having any sentence longer than 500 characters. We also rejected those with the number of tokens in any sentence higher than 100. Moreover, we applied FastText's LID-201 model~\cite{burchell-etal-2023-open} to verify whether the source and target sentences are indeed expressed in the correct language (the expected language should have probability score of at least 10\%).

For language pairs where English is present as a source or target language, we used Bicleaner AI package, which estimates the likelihood of a pair of sentences being mutual translations~\cite{zaragoza-bernabeu-etal-2022-bicleaner}. We set the bicleaner-score threshold to 50\%, rejecting everything below this score. Bicleaner cannot be applied to other languages as for now because it does not provide open-source models for such language pairs.



\paragraph{Step 2: Formality classification}
The next step involved extracting two subsets for each language pair: those with formal and informal translations, respectively. For this reason, for each target language, we searched for formality classifiers or golden standard annotations that can be used to train a classifier. As discussed in Section~\ref{sec:discussion}, we only need to classify the target language according to its formality level. There were three scenarios to address:
\begin{itemize}
    \item Formality classifier available: For English, there is a publicly available classifier available online~\footnote{\url{https://huggingface.co/s-nlp/roberta-base-formality-ranker}}, the quality of which is proven by an accompanying research paper~\cite{10.1007/978-3-031-35320-8_4}. As the classifier was pre-trained on the biggest dataset available (\textbf{GYAFC}) and produces the probability of a given example to be formal, we selected this classifier to process those pairs of sentences where the target language is English.
    \item No formality classifier, but golden standard dataset available: For German, French, Italian, Spanish, Portuguese, and Dutch, no formality classifier accompanied by a research paper could be found. For this reason, we decided to train a classifier using \textbf{CoCoA-MT} dataset, as it provides pairs of formal and informal forms of sentences for each of the aforementioned target languages.
    \item No formality classifier nor dataset available: For Polish, we did not even have a golden standard dataset to train a classifier on. For that language, we created a hold-out subset of sentences in Polish downloaded using \textbf{MTData}, and then the set was annotated by a group of six native speakers. This way, we collected examples of formal and informal sentences, which were of roughly the same size as golden standard examples provided in \textbf{CoCoA-MT} and were split into train and test sets to mimic the structure of \textbf{CoCoA-MT}.
\end{itemize}

We decided to include the Polish dataset in examples collected from \textbf{CoCoA-MT} and tried to fine-tune a single multilingual classifier to capture inter-lingual relations. 

However, early experiments with models described later in this Section showed that fine-tuning a pre-trained model using \textbf{CoCoA-MT} combined with Polish leads to extreme probabilities assigned to most sentences collected from \textbf{MTData}. As the dataset contains only formal and informal examples, classifiers learn to treat every sentence as either formal or informal, while in many cases (especially sentences that are not related to an interlocutor) they are neither formal nor informal. 

An analysis of annotations from \textbf{CoCoA-MT} showed that formality and informality are frequently expressed using appropriate personal pronouns. For this reason, we decided to generate an additional neutral dataset for each language generated as a random hold-out sample of target sentences from \textbf{MTData} that does not contain any phrase marked in \textbf{CoCoA-MT} as either formal or informal. For Polish, we continued the annotation task, asking natives to identify neutral examples in the hold-out dataset.

We fine-tuned several language models verifying how big a neutral sample size should be to maximize the scores. At each verification step, we changed only the size of the neutral set in the train set, while preserving a constant set of 600 neutral examples in the test set. We evaluated accuracy score for mDeberta-v3-base~\cite{he2021debertav3}, XLM-RoBERTa-uncased~\cite{DBLP:journals/corr/abs-1911-02116}, and BERT-base-multilingual-uncased~\cite{Devlin2019BERTPO}. The results, summarized in Table~\ref{tab:neutral_sample_size} show that the best average accuracy score is achieved using mDeberta-v3-base with a neutral training sample size $= 500$. This comes in line with the trainset size of original \textbf{CoCoA-MT}, as it uses 400 formal and 400 informal examples for each language pair, so including a neutral sample of this size results in an approximately balanced training dataset.

Diving deeper into per-language scores, as presented in Table~\ref{tab:per_language_scores}, we observe that all \textbf{CoCoA-MT} languages obtain very high accuracy scores. This may be due to the fact that frequently examples in \textbf{CoCoA-MT} distinguish formality based on personal pronouns. For Polish, the scores are lower as personal pronouns are often dropped and the form of a verb may be used to express a linguistic person implicitly. However, even for Polish, the scores are much higher than random guesses. For English, we report the scores for \textbf{GYAFC} obtained from a research paper that a given model accompanies~\cite{10.1007/978-3-031-35320-8_4}.


\begin{table}
\begin{center}
\begin{tabular}{|c|c|c|c|}
 \hline NSS & mDeberta & XLM-RoBERTa & mBERT \\ \hline 
 100 & 0.9304 & 0.9301 & 0.9297 \\ 
 200 & 0.9475 & 0.9437 & 0.9395 \\  
 300 & 0.9519 & 0.9465 & 0.9454 \\  
 400 & 0.9531 & 0.9511 & 0.9455 \\  
 500 & \textbf{0.9552} & 0.9523 & 0.9499 \\
 600 & 0.9551 & 0.9534 & 0.9492 \\ \hline
\end{tabular}
\caption{Neutral sample size (NSS) for training vs. average accuracy score (over all languages) of a given model. mDeberta stands for mDeberta-v3-base, XLM-RoBERTa stands for XLM-RoBERTa-uncased, and mBERT stands for BERT base multilingual uncased.}
\label{tab:neutral_sample_size}
\end{center}
\end{table}

\begin{table*}
\begin{center}
\begin{tabular}{|r|l|l|l|l|l|l||l|}
\hline Language & German & French & Italian & Dutch & Polish & Portuguese & English \\
\hline Accuracy & 0.9928 & 0.9926 & 0.9772 & 0.9962 & 0.7861 & 0.9789 & 0.9~\cite{10.1007/978-3-031-35320-8_4} \\
 \hline 
\end{tabular}
\caption{Classification scores for each language using mDeberta-v3-base and neutral sample size = 500 in comparison to English classifier trained on \textbf{GYAFC}. The average score of mDeberta-v3-base is equal to 0.9552.}
\label{tab:per_language_scores}
\end{center}
\end{table*}

\paragraph{Step 3: FAME-MT compilation}
The dataset compilation process was performed as follows: For each language $src$ from source languages set and each target language $tgt$, we classified all targets among translations between $src$ and $tgt$. When the target language was English, we used a model trained on \textbf{GYAFC}~\cite{10.1007/978-3-031-35320-8_4}. For other languages, we used mDeberta-v3-base, which we fine-tuned in the previous step. Since mDeberta-v3-base returns probability distribution over three possible classes: formal, informal, and neutral, we chose the class with the highest probability as the final model decision. However, since the model trained on \textbf{GYAFC} returns only the probabilities of formal and informal classes (that always sum up to one), we split the probability range into three equal parts, treating examples assigned with formal class probability in ranges: $<0, \frac{1}{3}>, (\frac{1}{3}, \frac{2}{3}>, (\frac{2}{3}, 1>$ as informal, neutral, and formal, respectively.


The classification process continued until we reached 50,000 informal and 50,000 formal examples for each language pair. Finally, for each language pair, we stored translations where the target sentence was considered formal or informal into separate files. 


The dataset is published online\footnote{\url{https://github.com/laniqo-public/fame-mt/}} along with the scripts used for analysis, formality classifiers, and MT models. It has the following structure: Each target language has its own folder assigned, and inside these folders, there is a separate folder for each source and target language combination. In those folders, there is a pair of files: \texttt{informal.tsv} and \texttt{formal.tsv} each providing 50,000 translations between source and target languages, where the target was considered informal and formal, respectively.

\section{Explorative analysis}

\begin{figure*}[ht]
  \includegraphics[width=\textwidth]{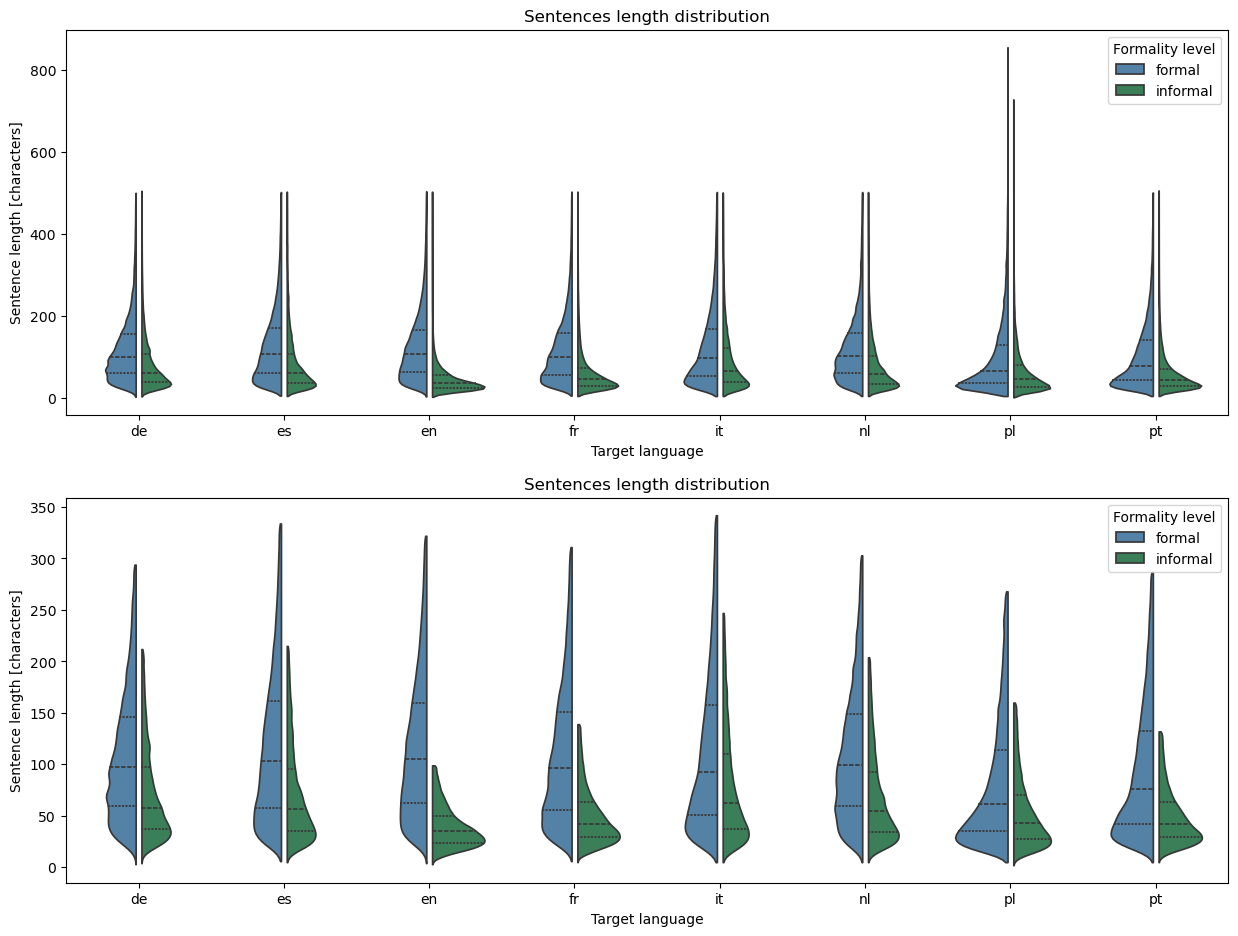}
  \caption{Violin plots representing the distributions of sentence lengths interpreted as the number of characters. The upper figure represents the distributions calculated over the original dataset. As it shows that there are some outliers with big values, we provide the lower figure generated over a subset of texts whose lengths are between Q1 - 1.5 IQR and Q3 + 1.5 IQR (Q1=first quartile, Q3=third quartile, IQR=inter-quartile range) to focus more on the most common scenarios.}
  \label{fig:sentence_length}
\end{figure*}

To verify the quality of the dataset, we selected several approaches that would show key characteristics of the dataset.

\paragraph{Average length of sentences} One of the basic metrics that can describe the difference between formal and  informal language is the average sentence length in each of the categories. We may expect that formal sentences may be longer than informal, which are frequently used to share knowledge quickly. Figure~\ref{fig:sentence_length} proves that for each target language, the formal translations are much longer than informal ones in the collected dataset.

\paragraph{Per-category key tokens}
To check which tokens characterize a given class, we aimed at understanding classifier decisions and feature importance. As the classifiers we used in the dataset creation process were too big to be handled by the popular explainable AI framework LIME~\cite{ribeiro2016should}, we decided to fit a lightweight linear model to \textbf{FAME-MT} and interpret its features to understand the difference between classes.

Having a set of sentences classified as formal and informal for each target language, we applied the TF-IDF vectorizer to a class-balanced sample of target sentences to identify the 200,000 most important tokens among target texts. Then, we used those tokens as features for a logistic regression model with the intercept value set to 0. This way, we ensure that the coefficients assigned by the logistic regression model correlate with a given class. As we modeled formal sentences as a positive class and informal ones as negative, tokens that are assigned with high positive coefficient values are strongly related to the formal class, and those associated with the highly negative values are strongly related to the informal class. The coefficients selected during training were then sorted for each language, and the top 100 tokens and bottom 100 were selected as those describing formal and informal categories, respectively.

Table~\ref{tab:log_reg} presents the most important features for each category and each language considered. As can be seen, for languages with formal \textit{you}, the most important words are personal pronouns (e.g., formal \textit{Sie} vs. informal \textit{du} in German or formal \textit{usted} and informal \textit{tu} in Spanish). For languages, where personal pronouns may be omitted in informal scenarios, the grammar form of verbs may indicate informality (e.g., \textit{jesteś} (informal \textit{you are}) in Polish). For English, where there is no formal \textit{you}, more sophisticated words are observed as the most important formal tokens, and contractions or swear words are found among the most characteristic informal tokens.

\begin{table*}
\small
\begin{center}
\begin{tabular}{|r|l|l|}
 \hline Language & formal words & informal words  \\ \hline 
 German (DE) & sie, ihre, ihr, ihnen, ihres, \ldots & du, deine, dir, dich, dein, \ldots \\ 
 English (EN) & distinctive, relations, moreover, obtain, refers, \ldots & gotta, f*****g, gonna, ain, wanna, \ldots \\
 Spanish (ES) & su, sus, usted, le, está, \ldots & tu, te, tus, estás, quieres, \ldots \\ 
 French (FR) & vous, votre, vos,  pouvez, avez, \ldots & tu, toi, te, ton, ta, \ldots \\ 
 Italian (IT) & sua, suo, suoi, lei, le, voi, \ldots & ti, hai, tuo, tua, tuoi, sai, tu, \ldots \\ 
 Dutch (NL) & uw, kunt, wilt, bent, hebt \ldots & je, jij, jullie, jouw, jou, \ldots \\ 
  Polish (PL) & pan, pani, wam, państwa, państwo, \ldots & jesteś, ci, możesz, myślisz, musisz, \ldots \\ 
 Portuguese (PT) & você, sua, seu, suas, seus, \ldots & te, teu, tua, tu, tens,\ldots \\  
 \hline

\end{tabular}
\caption{Words with the strongest relation to classes selected by analyzing feature coefficients in logistic regression model fitted on \textbf{FAME-MT}.}
\label{tab:log_reg}
\end{center}
\end{table*}

\paragraph{Textual complexity} We used textacy\footnote{\url{https://pypi.org/project/textacy/}} to measure the complexity of target sentences classified as formal and informal. The motivation for this step was the intuition that informal language may be easier to comprehend while formal texts should be harder. For English, we used the automated readability index function, which calculates the relation between the number of characters in a given sequence related to the number of words and sentences and is calculated using the following formula: $4.71 \cdot count_{chars} / count_{words} + 0.5 \cdot count_{words} / count_{sents} - 21.43$. We calculated the score for all language pairs with English as a target and calculated the mean value. We obtained $11.54$ for formal sentences and $4.28$ for informal ones, which means that formal sentences are more complex than informal ones. Considering each language pair with the target set to English, formal translations are more complex every time. The biggest difference between formal and informal sentence scores is seen in the case of French-English (formal: 12.856, informal: 2.788) and the smallest in the case of Czech and English (formal: 11.175, informal: 7.005). We analyzed the Flesch score, which indicates the reading ease for sentences collected. We sampled 5,000 sentences for each source - target language pair, where target sentence was among  languages supported by textacy (German, English, French, Italian, Spanish, Portuguese, Dutch). We present the full analysis in Table~\ref{tab:flesch} provided in Appendix A. For each language pair, the informal texts are scored higher than formal ones, which indicates that informal texts are easier to read. This follows the general intuition. Averaging over all source languages, the average Flesch scores for formal and informal texts are (F and I represent formal and informal, respectively): German: 51.99 F and 72.57 I, English: 63.19 F \& 90.89 I, Spanish: 68.93 F \& 80.45 I, French: 78.33 F \& 99.69 I, Italian: 77.65 F \& 81.4 I, Dutch: 57.61 F \& 76.32 I, Portuguese: 66.9 F \& 76.32 I.

\paragraph{Source vs. target sentence formality}
To explore the relation between the formality level of source sentence and its translation in target language, we selected a subset of \textbf{FAME-MT} for which we had classifiers for both source and target languages. As these are eight languages, from all 112 language pairs, we selected 56 of them (8 times 8 possible language pairs minus 8 pairs where both source and target are the same). Then, for each target language considered, we analyzed formal translations and informal translations separately: for each formal translation, we used appropriate classifier to verify whether a given source sentence is also formal and for each informal translation, we used appropriate classifier to verify whether a given source sentence is also informal. Then, we calculated the percentage of cases where the agreement was observed. As presented in Table~\ref{tab:cross_language_formality}, on average, 38.81\% of examples for which target sentence was considered formal had the source sentence classified as formal. Analogously, 40.28\% of examples classified as informal had the source sentence classified as informal. These results show that the formality level of the source does not determine the formality level of the target text. However, there is some correlation as the scores are higher than a random choice, which would achieve 33\% (assigning a random class from: formal, informal, neutral). Diving into language pairs as presented in Table~\ref{tab:confusion_matrix} in Appendix A and considering translations that are formal in target language, the strongest relation is between French sources and Dutch targets (82.8\% of formal cases in Dutch are also formal in French), and the weakest between Polish sources and English targets (5.6\% of formal cases in English are also formal in Polish). Considering translations that are informal in target language, the strongest relation is between Polish sources and French targets (73.4\% of informal cases in French are also informal in Polish), and the weakest between German sources and English targets (5.6\% of informal cases in English are also informal in German).

\begin{table}
\begin{center}
\begin{tabular}{|r|c|c|}
 \hline Language & Formal S+T & Informal S+T\\ \hline 
 German (DE) & 42.82\% & 48.41\% \\
 English (EN) & 17.71\% & 14.95\% \\
 Spanish (ES) & 41.18\% & 40.82\% \\ 
 French (FR) & 45.51\% & 54.4\% \\ 
 Italian (IT) & 37.91\% & 37.36\%\\ 
 Dutch (NL) & 55.56\% & 39.38\% \\ 
 Polish (PL) & 37.16\% & 30.87\% \\ 
 Portuguese (PT) & 35.67\% & 56.04\% \\ \hline
 \textbf{Average} & 38.81\% & 40.28\%\\ 
 \hline

\end{tabular}
\caption{Percentage of examples where formal source co-occurs with formal target (Formal S+T), and percentage of examples, where informal source co-occurs with informal target (Informal S+T). Formality of source and targets was determined using classifiers. The row \textit{Average} represents the micro-average over all languages.}
\label{tab:cross_language_formality}
\end{center}
\end{table}

\section{Application to Machine Translation}

This section explores the application of the \textbf{FAME-MT} dataset to enable control over the formality tone of translated text. We conducted experiments on two translation directions: English-German and English-Polish, leveraging pre-trained models from the OPUS collection~\cite{tiedemann-2020-tatoeba} for fine-tuning.

\paragraph{Fine-tuning pre-trained MT models}
To facilitate the fine-tuning process, we augmented the existing model vocabulary with two new tokens: \texttt{<FORMAL>} and \texttt{<INFORMAL>}. Additionally, we expanded the existing model embeddings to accommodate these new vocabulary items by initializing them with zeros.

To maintain consistency with the baseline models, we reused their Transformer-based~\cite{attention-all-you-need} architecture and training hyperparameters, except for batch size and learning rate. The English-German direction utilizes the Transformer-big architecture, while the English-Polish direction uses the Transformer-base architecture due to the lack of a larger model in the OPUS collection for this translation direction. Both fine-tuned models employed a small batch size of 5,000 target tokens and a learning rate of 1e-4. As the original models were trained using the Marian~\cite{mariannmt} framework, we utilized it for fine-tuning as well.

We pre-processed the \textbf{FAME-MT} dataset with pre-trained SentencePiece~\cite{kudo-richardson-2018-sentencepiece} tokenizers included in the model package, and then prepended \texttt{<FORMAL>} and \texttt{<INFORMAL>} tags to the formal and informal parts of the source input, respectively. To guarantee a diverse validation set during training, we randomly sampled 500 sentences from each formality category and added 500 random neutral samples not using personal pronouns, resulting in a total of 1500 validation samples per translation direction. The remaining data was used for training.

\paragraph{Automatic evaluation and contrastive examples}
Three decoding modes were evaluated for the fine-tuned models: standard, formal tone, and informal tone. Their performance was compared to the baseline OPUS model on the Flores (devtest)~\cite{nllb-22} and NTREX~\cite{federmann-etal-2022-ntrex} datasets, using BLEU~\cite{papineni-etal-2002-bleu}, chrF~\cite{popovic-2015-chrf}, and COMET\footnote{\texttt{wmt22-comet-da} COMET model was used}~\cite{comet} metrics for evaluation. The chrF and BLEU results were computed with sacreBLEU\footnote{
BLEU signature: nrefs:1|case:mixed|eff:no|tok:13a\\|smooth:exp|version:2.3.1}\footnote{chrF signature: nrefs:1|case:mixed|eff:yes|nc:6|nw:0\\|space:no|version:2.3.1}~\cite{post-2018-call}. The evaluation results are presented in Tables~\ref{tab:mt-eval-flores} and ~\ref{tab:mt-eval-ntrex} for Flores and NTREX datasets, respectively.

The results for English-German translation were encouraging. Fine-tuning on \textbf{FAME-MT} did not negatively impact overall translation quality, and automatic metrics even suggested a slight improvement. However, the English-Polish model exhibited a slight decrease in quality, potentially due to the smaller model size being more susceptible to the impact of embedding extension. 

Tables~\ref{tab:mt-examples-polish} and~\ref{tab:mt-examples-german} showcase various translations of the same sentence with different formality levels for each fine-tuned model. Interestingly, in languages like Polish, where gender neutrality is not present in singular pronouns, formal translations can introduce gender bias. For example, English \textit{you} can be translated as either \textit{Pan} (masculine) or \textit{Pani} (feminine). In such cases, sentence-level machine translation might struggle to choose the correct form without additional context. As expected, formal translations generally tend to be more literal compared to their informal counterparts.

\paragraph{Released models}
To promote further research and facilitate accessibility, we have made the fine-tuned models publicly available as open-source resources~\footnote{\url{https://github.com/laniqo-public/fame-mt/}}.

This work demonstrates the potential of incorporating formality control datasets into machine translation pipelines. While further investigation is needed to refine the approach for different model sizes and language pairs, the results show promise for producing translations that accurately reflect the desired level of formality. The open-sourced models will enable  researchers to build upon this work and explore applications in diverse scenarios.

\begin{table*}[]
\begin{center}
\begin{tabular}{clrrrrrr}
\hline
\multicolumn{2}{c}{\multirow{3}{*}{\textbf{Model}}}                 & \multicolumn{6}{c}{\textbf{Flores}}                                                                                                                                                                                       \\ \cline{3-8} 
\multicolumn{2}{c}{}                                                & \multicolumn{3}{c|}{English $\rightarrow$ German}                                                            & \multicolumn{3}{c}{English $\rightarrow$ Polish}                                                           \\ \cline{3-8} 
\multicolumn{2}{c}{}                                                & \multicolumn{1}{c}{\textbf{COMET}} & \multicolumn{1}{c}{\textbf{chrF}} & \multicolumn{1}{c|}{\textbf{BLEU}}  & \multicolumn{1}{c}{\textbf{COMET}} & \multicolumn{1}{c}{\textbf{chrF}} & \multicolumn{1}{c}{\textbf{BLEU}} \\ \hline
\multicolumn{1}{l|}{\multirow{3}{*}{OPUS-Finetuned-50k}} & Standard & \textbf{0.8687}                    & 66.14                             & \multicolumn{1}{r|}{39.98}          & 0.8566                             & 50.54                             & 19.87                             \\
\multicolumn{1}{l|}{}                                    & Formal   & 0.8656                             & 66.18                             & \multicolumn{1}{r|}{\textbf{40.02}} & 0.8523                             & 50.47                             & 19.70                             \\
\multicolumn{1}{l|}{}                                    & Informal & \textbf{0.8687}                    & 65.85                             & \multicolumn{1}{r|}{39.60}          & 0.8539                             & 50.48                             & 19.84                             \\ \hline
OPUS (baseline)                                          & Standard & 0.8675                             & \textbf{66.20}                    & \multicolumn{1}{r|}{39.90}          & \textbf{0.8624}                    & \textbf{50.85}                    & \textbf{20.15}   \\ \hline                
\end{tabular}
\caption{Automatic evaluation results on the Flores dataset before and after fine-tuning the model on the FAME-MT dataset.}
\label{tab:mt-eval-flores}
\end{center}
\end{table*}

\begin{table*}[]
\begin{center}
\begin{tabular}{clrrrrrr}
\hline
\multicolumn{2}{c}{\multirow{3}{*}{\textbf{Model}}}                 & \multicolumn{6}{c}{\textbf{NTREX}}                                                                                                                                                                                        \\ \cline{3-8} 
\multicolumn{2}{c}{}                                                & \multicolumn{3}{c|}{English $\rightarrow$ German}                                                            & \multicolumn{3}{c}{English $\rightarrow$ Polish}                                                           \\ \cline{3-8} 
\multicolumn{2}{c}{}                                                & \multicolumn{1}{c}{\textbf{COMET}} & \multicolumn{1}{c}{\textbf{chrF}} & \multicolumn{1}{c|}{\textbf{BLEU}}  & \multicolumn{1}{c}{\textbf{COMET}} & \multicolumn{1}{c}{\textbf{chrF}} & \multicolumn{1}{c}{\textbf{BLEU}} \\ \hline
\multicolumn{1}{l|}{\multirow{3}{*}{OPUS-Finetuned-50k}} & Standard & 0.8295                             & \textbf{60.25}                    & \multicolumn{1}{r|}{\textbf{32.23}} & 0.8097                             & 51.45                             & 23.03                             \\
\multicolumn{1}{l|}{}                                    & Formal   & 0.8256                             & 60.21                             & \multicolumn{1}{r|}{31.98}          & 0.8049                             & 51.32                             & 22.83                             \\
\multicolumn{1}{l|}{}                                    & Informal & \textbf{0.8328}                    & 60.15                             & \multicolumn{1}{r|}{31.95}          & 0.8047                             & 51.22                             & 22.77                             \\ \hline
OPUS (baseline)                                          & Standard & 0.8251                             & 59.91                             & \multicolumn{1}{r|}{31.64}          & \textbf{0.8125}                    & \textbf{51.89}                    & \textbf{23.41}  \\ \hline                 
\end{tabular}
\caption{Automatic evaluation results on the NTREX dataset before and after fine-tuning the model on the FAME-MT dataset.}
\label{tab:mt-eval-ntrex}
\end{center}
\end{table*}

\begin{table*}[]
\begin{center}
\begin{tabular}{p{5cm}p{5cm}p{5cm}}
\hline
\multicolumn{1}{c}{\textbf{English}} &
  \multicolumn{1}{c}{\textbf{Polish (Formal)}} &
  \multicolumn{1}{c}{\textbf{Polish (Informal)}} \\ \hline
\textcolor{ForestGreen}{You} have to tip your cap. &
  \textcolor{ForestGreen}{Musi pan} przechylić czapkę. &
  \textcolor{ForestGreen}{Musisz} przechylić czapkę. \\ \hline
They think \textcolor{ForestGreen}{you're} sad and will be pleased because they got to you. &
  Sądzą, że \textcolor{ForestGreen}{jest pan} smutny i będą zadowoleni, bo do pana doszli. &
  Sądzą, że \textcolor{ForestGreen}{jesteś} smutny i będą zadowoleni, bo cię dorwali. \\ \hline
They don't know \textcolor{ForestGreen}{you're} furious. &
  Nie wiedzą, że \textcolor{ForestGreen}{jest pani} wściekła. &
  Nie wiedzą, że \textcolor{ForestGreen}{jesteś} wściekła. \\ \hline
But on Saturday, North Korean Foreign Minister Ri Yong-ho \textcolor{ForestGreen}{blamed} US sanctions for the lack of progress since then. &
  Ale w sobotę, północnokoreański minister spraw zagranicznych Ri Yong-ho \textcolor{ForestGreen}{obarczył} sankcje USA za brak postępów od tego czasu. &
  Ale w sobotę, północnokoreański minister spraw zagranicznych Ri Yong-ho \textcolor{ForestGreen}{obwiniał} USA o sankcje za brak postępów od tego czasu. \\ \hline
Ring also settled a lawsuit with competing \textcolor{ForestGreen}{security company}, the ADT Corporation. &
  Ring rozstrzygnął również sprawę z konkurencyjnym \textcolor{ForestGreen}{koncernem ochroniarskim}, korporacją ADT. &
  Ring rozstrzygnął również sprawę z konkurencyjną \textcolor{ForestGreen}{firmą ochroniarską}, firmą ADT Corporation. \\ \hline
\end{tabular}%
\caption{Examples of different machine translation results with formal/informal tone in English $\rightarrow$ Polish translation.}
\label{tab:mt-examples-polish}
\end{center}
\end{table*}

\begin{table*}[]
\begin{center}
\begin{tabular}{p{5cm}p{5cm}p{5cm}}
\hline
\multicolumn{1}{c}{\textbf{English}} &
  \multicolumn{1}{c}{\textbf{German (Formal)}} &
  \multicolumn{1}{c}{\textbf{German (Informal)}} \\ \hline
  \textcolor{ForestGreen}{You} just \textcolor{ForestGreen}{have to} pay them the right amount of respect, he said.. &
  \textcolor{ForestGreen}{Du musst} ihnen nur den richtigen Respekt erweisen, sagte er. &
  \textcolor{ForestGreen}{Sie müssen} ihnen nur die richtige Menge an Respekt zahlen, sagte er. \\ \hline
Sit down. &
  Setzen Sie sich. &
  Setz dich. \\ \hline
\textcolor{ForestGreen}{Would you like} to go with me to the cinema next week? &
 \textcolor{ForestGreen}{Möchtest du} nächste Woche mit mir ins Kino gehen? &
\textcolor{ForestGreen}{Möchten Sie} nächste Woche mit mir ins Kino gehen? \\ \hline
I'll be with you soon. &
Ich werde bald bei dir sein. &
Ich bin bald bei Ihnen. \\ \hline
\textcolor{ForestGreen}{I would like} a word with your boss. &
\textcolor{ForestGreen}{Ich hätte gerne} ein Wort mit \textcolor{ForestGreen}{deinem} Chef. &
\textcolor{ForestGreen}{Ich möchte} ein Wort mit \textcolor{ForestGreen}{Ihrem} Chef.\\ \hline
\end{tabular}%
\caption{Examples of different machine translation results with formal/informal tone in English $\rightarrow$ German translation.}
\label{tab:mt-examples-german}
\end{center}
\end{table*}

\section{Discussion}
\label{sec:discussion}

As we have shown in the previous section, the \textbf{FAME-MT} dataset can make pre-trained models formality-aware. Also, metrics used to analyze the dataset show that the characteristics of the dataset are consistent with intuition: formal documents are harder to read, they tend to be longer or use more formal words.

An interesting observation is that even though the \textbf{CoCoA-MT} dataset introduces a relatively small dataset of formal and informal examples per each language, the accuracy of the pre-trained mDeberta-v3-base model, which is fine-tuned on \textbf{CoCoA-MT} is very high when measured on \textbf{CoCoA-MT}'s test set with neutral samples added (0.9552). An analysis of the annotations provided in \textbf{CoCoA-MT} reveals that it focuses on expressing formality using personal pronouns. However, formality level, in general, can be expressed using various language constructs, e.g., contractions, formal greetings, slang words, and appropriate personal pronouns. On the other hand, e.g.,  \textbf{GYAFC} focuses more on constructs other than personal pronouns due to the lack of formal \textit{you} in English. We think that this is the reason for the high scores observed for a small dataset (\textbf{CoCoA-MT}) and lower scores for a much bigger dataset (\textbf{GYAFC}). The focus of \textbf{CoCoA-MT} on pronouns leads to too optimistic quality estimates as compared to the evaluation provided for the English classifier.

However, since formal constructs co-occur with each other, classifiers trained using \textbf{CoCoA-MT} work well in practice. As we can see in Tables~\ref{tab:mt-examples-polish} and~\ref{tab:mt-examples-german} -- models trained with \textbf{FAME-MT} introduce subtle differences depending on the expected formality level (e.g., \textit{hätte gerne} vs. \textit{möchte} or \textit{koncernem ochroniarskim} vs. \textit{firmą ochroniarską}).

The motivation for using classifiers to classify only sentences in target languages is supported by the analysis of source vs. target sentence formality. As we have shown that the formality level of the source language does not determine the formality of the target language, in \textbf{FAME-MT}, we collect sentences of various levels of formality mapped to a given target sentence formality level. This way, we can inject a token representing the desired formality level (be it \textit{formal} or \textit{informal}) to steer the expected formality level. Thus, having only target sentences classified, we can fine-tune machine translation models to become formality-aware, regardless of the formality level of the source sequence.

\section{Addressing research questions} In this paper, we stated three research questions that can be answered now. Regarding the first one, we show that using a set of classifiers for the parallel corporas' target languages, we created a large dataset, which is useful for training formality-aware MT models. We proved it by fine-tuning general machine translation models and utilizing metrics confirming intuitions, thus, the answer to \textbf{RQ1} is positive. Also, the experiments show that fine-tuning pre-trained OPUS models with 50,000 formal and 50,000 informal examples is enough for fine-tuning English-German and English-Polish pairs of languages. The quality of the English-German model is higher but the quality of English-Polish is also satisfactory. For this reason, we can give the positive answer to \textbf{RQ2}. Finally, the classifiers used reveal that the formality level of the source language does not determine the formality level of the target language in the datasets collected using \textbf{MTData}. This observation justifies the need to inject a special formality token to ensure a given formality level of the translation. Thus, the answer to \textbf{RQ3} is negative. 

\section{Conclusions}
In this paper, we introduce \textbf{FAME-MT} - a dataset consisting of 11.2 million translations between 112 European language pairs, where sentences in target languages are classified as formal or informal. As the dataset is a computer-generated silver standard, we used a set of metrics to prove the good quality of the data. Moreover, proof-of-concept models fine-tuned using formality data show that the dataset can be successfully utilized in problems requiring enforcing a given formality level of the system's output. Due to its size and large number of language pairs selected, \textbf{FAME-MT} is the largest and most diverse dataset available, introducing 18 times more European language pairs than biggest existing multilingual datasets (112 language pairs vs. 6 in \textbf{CoCoA-MT}) and providing over 100 times more sentences annotated with formality level than the biggest datasets (11.2 million vs. 110,000 in \textbf{GYAFC}). We made the dataset publicly accessible, and provided all the source codes for rerunning the analysis~\footnote{\url{https://github.com/laniqo-public/fame-mt/}}. We hope that this dataset may help to produce better formality-aware machine translation models especially for pairs of languages that were not yet covered or underrepresented in existing datasets (e.g., Czech $\rightarrow$ French, Polish $\rightarrow$ German, or Danish $\rightarrow$ Dutch).


\bibliography{eamt24}

\begin{thebibliography}{}

\bibitem[\protect\citename{Babakov \bgroup et al.\egroup }2023]{10.1007/978-3-031-35320-8_4}
Babakov, Nikolay, David Dale, Ilya Gusev, Irina Krotova, and Alexander Panchenko.
\newblock 2023.
\newblock Don't lose the message while paraphrasing: A study on content preserving style transfer.
\newblock In M{\'e}tais, Elisabeth, Farid Meziane, Vijayan Sugumaran, Warren Manning, and Stephan Reiff-Marganiec, editors, {\em Natural Language Processing and Information Systems}, pages 47--61, Cham. Springer Nature Switzerland.

\bibitem[\protect\citename{Ba{\~n}{\'o}n \bgroup et al.\egroup }2020]{banon-etal-2020-paracrawl}
Ba{\~n}{\'o}n, Marta, Pinzhen Chen, Barry Haddow, Kenneth Heafield, Hieu Hoang, Miquel Espl{\`a}-Gomis, Mikel~L. Forcada, Amir Kamran, Faheem Kirefu, Philipp Koehn, Sergio Ortiz~Rojas, Leopoldo Pla~Sempere, Gema Ram{\'\i}rez-S{\'a}nchez, Elsa Sarr{\'\i}as, Marek Strelec, Brian Thompson, William Waites, Dion Wiggins, and Jaume Zaragoza.
\newblock 2020.
\newblock {P}ara{C}rawl: Web-scale acquisition of parallel corpora.
\newblock In Jurafsky, Dan, Joyce Chai, Natalie Schluter, and Joel Tetreault, editors, {\em Proceedings of the 58th Annual Meeting of the Association for Computational Linguistics}, pages 4555--4567, Online, July. Association for Computational Linguistics.

\bibitem[\protect\citename{Biber and Conrad}2019]{biber2019register}
Biber, Douglas and Susan Conrad.
\newblock 2019.
\newblock {\em Register, genre, and style}.
\newblock Cambridge University Press.

\bibitem[\protect\citename{Briakou \bgroup et al.\egroup }2021]{DBLP:conf/naacl/BriakouLZT21}
Briakou, Eleftheria, Di~Lu, Ke~Zhang, and Joel~R. Tetreault.
\newblock 2021.
\newblock Ol{\'{a}}, bonjour, salve! {XFORMAL:} {A} benchmark for multilingual formality style transfer.
\newblock In Toutanova, Kristina, Anna Rumshisky, Luke Zettlemoyer, Dilek Hakkani{-}T{\"{u}}r, Iz~Beltagy, Steven Bethard, Ryan Cotterell, Tanmoy Chakraborty, and Yichao Zhou, editors, {\em Proceedings of the 2021 Conference of the North American Chapter of the Association for Computational Linguistics: Human Language Technologies, {NAACL-HLT} 2021, Online, June 6-11, 2021}, pages 3199--3216. Association for Computational Linguistics.

\bibitem[\protect\citename{Burchell \bgroup et al.\egroup }2023]{burchell-etal-2023-open}
Burchell, Laurie, Alexandra Birch, Nikolay Bogoychev, and Kenneth Heafield.
\newblock 2023.
\newblock An open dataset and model for language identification.
\newblock In Rogers, Anna, Jordan Boyd-Graber, and Naoaki Okazaki, editors, {\em Proceedings of the 61st Annual Meeting of the Association for Computational Linguistics (Volume 2: Short Papers)}, pages 865--879, Toronto, Canada, July. Association for Computational Linguistics.

\bibitem[\protect\citename{Conneau \bgroup et al.\egroup }2019]{DBLP:journals/corr/abs-1911-02116}
Conneau, Alexis, Kartikay Khandelwal, Naman Goyal, Vishrav Chaudhary, Guillaume Wenzek, Francisco Guzm{\'{a}}n, Edouard Grave, Myle Ott, Luke Zettlemoyer, and Veselin Stoyanov.
\newblock 2019.
\newblock Unsupervised cross-lingual representation learning at scale.
\newblock {\em CoRR}, abs/1911.02116.

\bibitem[\protect\citename{Costa-juss{\`a} \bgroup et al.\egroup }2022]{nllb-22}
Costa-juss{\`a}, Marta~R, James Cross, Onur {\c{C}}elebi, Maha Elbayad, Kenneth Heafield, Kevin Heffernan, Elahe Kalbassi, Janice Lam, Daniel Licht, Jean Maillard, et~al.
\newblock 2022.
\newblock No language left behind: Scaling human-centered machine translation.
\newblock {\em arXiv preprint arXiv:2207.04672}.

\bibitem[\protect\citename{Devlin \bgroup et al.\egroup }2019]{Devlin2019BERTPO}
Devlin, Jacob, Ming-Wei Chang, Kenton Lee, and Kristina Toutanova.
\newblock 2019.
\newblock Bert: Pre-training of deep bidirectional transformers for language understanding.
\newblock In {\em North American Chapter of the Association for Computational Linguistics}.

\bibitem[\protect\citename{Eder \bgroup et al.\egroup }2023]{DBLP:conf/eacl/EderKW23}
Eder, Elisabeth, Ulrike Krieg{-}Holz, and Michael Wiegand.
\newblock 2023.
\newblock A question of style: {A} dataset for analyzing formality on different levels.
\newblock In Vlachos, Andreas and Isabelle Augenstein, editors, {\em Findings of the Association for Computational Linguistics: {EACL} 2023, Dubrovnik, Croatia, May 2-6, 2023}, pages 568--581. Association for Computational Linguistics.

\bibitem[\protect\citename{Federmann \bgroup et al.\egroup }2022]{federmann-etal-2022-ntrex}
Federmann, Christian, Tom Kocmi, and Ying Xin.
\newblock 2022.
\newblock {NTREX}-128 {--} news test references for {MT} evaluation of 128 languages.
\newblock In {\em Proceedings of the First Workshop on Scaling Up Multilingual Evaluation}, pages 21--24, Online, nov. Association for Computational Linguistics.

\bibitem[\protect\citename{Feely \bgroup et al.\egroup }2019]{DBLP:conf/aclwat/FeelyHG19}
Feely, Weston, Eva Hasler, and Adri{\`{a}} de~Gispert.
\newblock 2019.
\newblock Controlling japanese honorifics in english-to-japanese neural machine translation.
\newblock In Nakazawa, Toshiaki, Chenchen Ding, Raj Dabre, Anoop Kunchukuttan, Nobushige Doi, Yusuke Oda, Ondrej Bojar, Shantipriya Parida, Isao Goto, and Hidaya Mino, editors, {\em Proceedings of the 6th Workshop on Asian Translation, WAT@EMNLP-IJCNLP 2019, Hong Kong, China, November 4, 2019}, pages 45--53. Association for Computational Linguistics.

\bibitem[\protect\citename{Garcia and Firat}2022]{DBLP:journals/corr/abs-2202-11822}
Garcia, Xavier and Orhan Firat.
\newblock 2022.
\newblock Using natural language prompts for machine translation.
\newblock {\em CoRR}, abs/2202.11822.

\bibitem[\protect\citename{Gowda \bgroup et al.\egroup }2021]{gowda-etal-2021-many}
Gowda, Thamme, Zhao Zhang, Chris Mattmann, and Jonathan May.
\newblock 2021.
\newblock Many-to-{E}nglish machine translation tools, data, and pretrained models.
\newblock In {\em Proceedings of the 59th Annual Meeting of the Association for Computational Linguistics and the 11th International Joint Conference on Natural Language Processing: System Demonstrations}, pages 306--316, Online, August. Association for Computational Linguistics.

\bibitem[\protect\citename{He \bgroup et al.\egroup }2021]{he2021debertav3}
He, Pengcheng, Jianfeng Gao, and Weizhu Chen.
\newblock 2021.
\newblock Debertav3: Improving deberta using electra-style pre-training with gradient-disentangled embedding sharing.

\bibitem[\protect\citename{Junczys-Dowmunt \bgroup et al.\egroup }2018]{mariannmt}
Junczys-Dowmunt, Marcin, Roman Grundkiewicz, Tomasz Dwojak, Hieu Hoang, Kenneth Heafield, Tom Neckermann, Frank Seide, Ulrich Germann, Alham Fikri~Aji, Nikolay Bogoychev, Andr\'{e} F.~T. Martins, and Alexandra Birch.
\newblock 2018.
\newblock Marian: Fast neural machine translation in {C++}.
\newblock In {\em Proceedings of ACL 2018, System Demonstrations}, pages 116--121, Melbourne, Australia, July. Association for Computational Linguistics.

\bibitem[\protect\citename{Koehn}2005]{koehn-2005-europarl}
Koehn, Philipp.
\newblock 2005.
\newblock {E}uroparl: A parallel corpus for statistical machine translation.
\newblock In {\em Proceedings of Machine Translation Summit X: Papers}, pages 79--86, Phuket, Thailand, September 13-15.

\bibitem[\protect\citename{Kudo and Richardson}2018]{kudo-richardson-2018-sentencepiece}
Kudo, Taku and John Richardson.
\newblock 2018.
\newblock {S}entence{P}iece: A simple and language independent subword tokenizer and detokenizer for neural text processing.
\newblock In Blanco, Eduardo and Wei Lu, editors, {\em Proceedings of the 2018 Conference on Empirical Methods in Natural Language Processing: System Demonstrations}, pages 66--71, Brussels, Belgium, November. Association for Computational Linguistics.

\bibitem[\protect\citename{Nadejde \bgroup et al.\egroup }2022]{DBLP:conf/naacl/NadejdeCHNFD22}
Nadejde, Maria, Anna Currey, Benjamin Hsu, Xing Niu, Marcello Federico, and Georgiana Dinu.
\newblock 2022.
\newblock Cocoa-mt: {A} dataset and benchmark for contrastive controlled {MT} with application to formality.
\newblock In Carpuat, Marine, Marie{-}Catherine de~Marneffe, and Iv{\'{a}}n Vladimir~Meza Ru{\'{\i}}z, editors, {\em Findings of the Association for Computational Linguistics: {NAACL} 2022, Seattle, WA, United States, July 10-15, 2022}, pages 616--632. Association for Computational Linguistics.

\bibitem[\protect\citename{Niu and Carpuat}2020]{DBLP:conf/aaai/NiuC20}
Niu, Xing and Marine Carpuat.
\newblock 2020.
\newblock Controlling neural machine translation formality with synthetic supervision.
\newblock In {\em The Thirty-Fourth {AAAI} Conference on Artificial Intelligence, {AAAI} 2020, The Thirty-Second Innovative Applications of Artificial Intelligence Conference, {IAAI} 2020, The Tenth {AAAI} Symposium on Educational Advances in Artificial Intelligence, {EAAI} 2020, New York, NY, USA, February 7-12, 2020}, pages 8568--8575. {AAAI} Press.

\bibitem[\protect\citename{Niu \bgroup et al.\egroup }2017]{DBLP:conf/emnlp/NiuMC17}
Niu, Xing, Marianna~J. Martindale, and Marine Carpuat.
\newblock 2017.
\newblock A study of style in machine translation: Controlling the formality of machine translation output.
\newblock In Palmer, Martha, Rebecca Hwa, and Sebastian Riedel, editors, {\em Proceedings of the 2017 Conference on Empirical Methods in Natural Language Processing, {EMNLP} 2017, Copenhagen, Denmark, September 9-11, 2017}, pages 2814--2819. Association for Computational Linguistics.

\bibitem[\protect\citename{Papineni \bgroup et al.\egroup }2002]{papineni-etal-2002-bleu}
Papineni, Kishore, Salim Roukos, Todd Ward, and Wei-Jing Zhu.
\newblock 2002.
\newblock {B}leu: a method for automatic evaluation of machine translation.
\newblock In {\em Proceedings of the 40th Annual Meeting of the Association for Computational Linguistics}, pages 311--318, Philadelphia, Pennsylvania, USA, July. Association for Computational Linguistics.

\bibitem[\protect\citename{Pavlick and Tetreault}2016]{pavlick-tetreault-2016-empirical}
Pavlick, Ellie and Joel Tetreault.
\newblock 2016.
\newblock An empirical analysis of formality in online communication.
\newblock {\em Transactions of the Association for Computational Linguistics}, 4:61--74.

\bibitem[\protect\citename{Popovi{\'c}}2015]{popovic-2015-chrf}
Popovi{\'c}, Maja.
\newblock 2015.
\newblock chr{F}: character n-gram {F}-score for automatic {MT} evaluation.
\newblock In {\em Proceedings of the Tenth Workshop on Statistical Machine Translation}, pages 392--395, Lisbon, Portugal, September. Association for Computational Linguistics.

\bibitem[\protect\citename{Post}2018]{post-2018-call}
Post, Matt.
\newblock 2018.
\newblock A call for clarity in reporting {BLEU} scores.
\newblock In {\em Proceedings of the Third Conference on Machine Translation: Research Papers}, pages 186--191, Brussels, Belgium, October. Association for Computational Linguistics.

\bibitem[\protect\citename{Rao and Tetreault}2018]{DBLP:conf/naacl/RaoT18}
Rao, Sudha and Joel~R. Tetreault.
\newblock 2018.
\newblock Dear sir or madam, may {I} introduce the {GYAFC} dataset: Corpus, benchmarks and metrics for formality style transfer.
\newblock In Walker, Marilyn~A., Heng Ji, and Amanda Stent, editors, {\em Proceedings of the 2018 Conference of the North American Chapter of the Association for Computational Linguistics: Human Language Technologies, {NAACL-HLT} 2018, New Orleans, Louisiana, USA, June 1-6, 2018, Volume 1 (Long Papers)}, pages 129--140. Association for Computational Linguistics.

\bibitem[\protect\citename{Rei \bgroup et al.\egroup }2020]{comet}
Rei, Ricardo, Craig Stewart, Ana~C Farinha, and Alon Lavie.
\newblock 2020.
\newblock {COMET}: A neural framework for {MT} evaluation.
\newblock In {\em Proceedings of the 2020 Conference on Empirical Methods in Natural Language Processing (EMNLP)}, pages 2685--2702, Online, November. Association for Computational Linguistics.

\bibitem[\protect\citename{Ribeiro \bgroup et al.\egroup }2016]{ribeiro2016should}
Ribeiro, Marco~Tulio, Sameer Singh, and Carlos Guestrin.
\newblock 2016.
\newblock " why should i trust you?" explaining the predictions of any classifier.
\newblock In {\em Proceedings of the 22nd ACM SIGKDD international conference on knowledge discovery and data mining}, pages 1135--1144.

\bibitem[\protect\citename{Schioppa \bgroup et al.\egroup }2021]{DBLP:conf/emnlp/SchioppaVSF21}
Schioppa, Andrea, David Vilar, Artem Sokolov, and Katja Filippova.
\newblock 2021.
\newblock Controlling machine translation for multiple attributes with additive interventions.
\newblock In Moens, Marie{-}Francine, Xuanjing Huang, Lucia Specia, and Scott~Wen{-}tau Yih, editors, {\em Proceedings of the 2021 Conference on Empirical Methods in Natural Language Processing, {EMNLP} 2021, Virtual Event / Punta Cana, Dominican Republic, 7-11 November, 2021}, pages 6676--6696. Association for Computational Linguistics.

\bibitem[\protect\citename{Sennrich \bgroup et al.\egroup }2016]{DBLP:conf/naacl/SennrichHB16}
Sennrich, Rico, Barry Haddow, and Alexandra Birch.
\newblock 2016.
\newblock Controlling politeness in neural machine translation via side constraints.
\newblock In Knight, Kevin, Ani Nenkova, and Owen Rambow, editors, {\em {NAACL} {HLT} 2016, The 2016 Conference of the North American Chapter of the Association for Computational Linguistics: Human Language Technologies, San Diego California, USA, June 12-17, 2016}, pages 35--40. The Association for Computational Linguistics.

\bibitem[\protect\citename{Tiedemann}2016]{tiedemann-2016-opus}
Tiedemann, J{\"o}rg.
\newblock 2016.
\newblock {OPUS} {--} parallel corpora for everyone.
\newblock In {\em Proceedings of the 19th Annual Conference of the European Association for Machine Translation: Projects/Products}, Riga, Latvia, May 30{--}June 1. Baltic Journal of Modern Computing.

\bibitem[\protect\citename{Tiedemann}2020]{tiedemann-2020-tatoeba}
Tiedemann, J{\"o}rg.
\newblock 2020.
\newblock The {T}atoeba {T}ranslation {C}hallenge {--} {R}ealistic data sets for low resource and multilingual {MT}.
\newblock In {\em Proceedings of the Fifth Conference on Machine Translation}, pages 1174--1182, Online, November. Association for Computational Linguistics.

\bibitem[\protect\citename{Vaswani \bgroup et al.\egroup }2017]{attention-all-you-need}
Vaswani, Ashish, Noam Shazeer, Niki Parmar, Jakob Uszkoreit, Llion Jones, Aidan~N Gomez, \L~ukasz Kaiser, and Illia Polosukhin.
\newblock 2017.
\newblock Attention is all you need.
\newblock In Guyon, I., U.~Von Luxburg, S.~Bengio, H.~Wallach, R.~Fergus, S.~Vishwanathan, and R.~Garnett, editors, {\em Advances in Neural Information Processing Systems}, volume~30. Curran Associates, Inc.

\bibitem[\protect\citename{Xue \bgroup et al.\egroup }2021]{DBLP:conf/naacl/XueCRKASBR21}
Xue, Linting, Noah Constant, Adam Roberts, Mihir Kale, Rami Al{-}Rfou, Aditya Siddhant, Aditya Barua, and Colin Raffel.
\newblock 2021.
\newblock mt5: {A} massively multilingual pre-trained text-to-text transformer.
\newblock In Toutanova, Kristina, Anna Rumshisky, Luke Zettlemoyer, Dilek Hakkani{-}T{\"{u}}r, Iz~Beltagy, Steven Bethard, Ryan Cotterell, Tanmoy Chakraborty, and Yichao Zhou, editors, {\em Proceedings of the 2021 Conference of the North American Chapter of the Association for Computational Linguistics: Human Language Technologies, {NAACL-HLT} 2021, Online, June 6-11, 2021}, pages 483--498. Association for Computational Linguistics.

\bibitem[\protect\citename{Yang and Klein}2021]{DBLP:conf/naacl/YangK21}
Yang, Kevin and Dan Klein.
\newblock 2021.
\newblock {FUDGE:} controlled text generation with future discriminators.
\newblock In Toutanova, Kristina, Anna Rumshisky, Luke Zettlemoyer, Dilek Hakkani{-}T{\"{u}}r, Iz~Beltagy, Steven Bethard, Ryan Cotterell, Tanmoy Chakraborty, and Yichao Zhou, editors, {\em Proceedings of the 2021 Conference of the North American Chapter of the Association for Computational Linguistics: Human Language Technologies, {NAACL-HLT} 2021, Online, June 6-11, 2021}, pages 3511--3535. Association for Computational Linguistics.

\bibitem[\protect\citename{Zaragoza-Bernabeu \bgroup et al.\egroup }2022]{zaragoza-bernabeu-etal-2022-bicleaner}
Zaragoza-Bernabeu, Jaume, Gema Ram{\'\i}rez-S{\'a}nchez, Marta Ba{\~n}{\'o}n, and Sergio Ortiz~Rojas.
\newblock 2022.
\newblock Bicleaner {AI}: Bicleaner goes neural.
\newblock In {\em Proceedings of the Thirteenth Language Resources and Evaluation Conference}, pages 824--831, Marseille, France, June. European Language Resources Association.

\end{thebibliography}
\bibliographystyle{eamt24}

\appendix
\section{Detailed analysis of the FAME-MT dataset}

To understand the \textbf{FAME-MT} dataset deeper, we include the extension of the explorative analysis by adding figures and tables that may help understand the dataset better. 

Table~\ref{tab:mt_fame} provides information about the size of each language pair subset in \textbf{FAME-MT}. For each language pair, 50\% of examples (50,000) are examples with formal targets and 50\% of examples with informal ones.

Table~\ref{tab:flesch} extends the discussion on reading scores in terms of Flesch readability scores. Here, for each language pair, we see that informal translations are easier to read than formal ones.

Table~\ref{tab:confusion_matrix} provides an in-depth analysis of the formality of source sentences in relation to the formality of target sentences extending information from Table~\ref{tab:cross_language_formality}. Here, we provide statistics for each language pair considering all kinds of disagreements separately, counting how many source sentences were classified as formal/informal/neutral when target sentence is formal or informal.

\begin{figure*}[ht]
  \includegraphics[width=\textwidth]{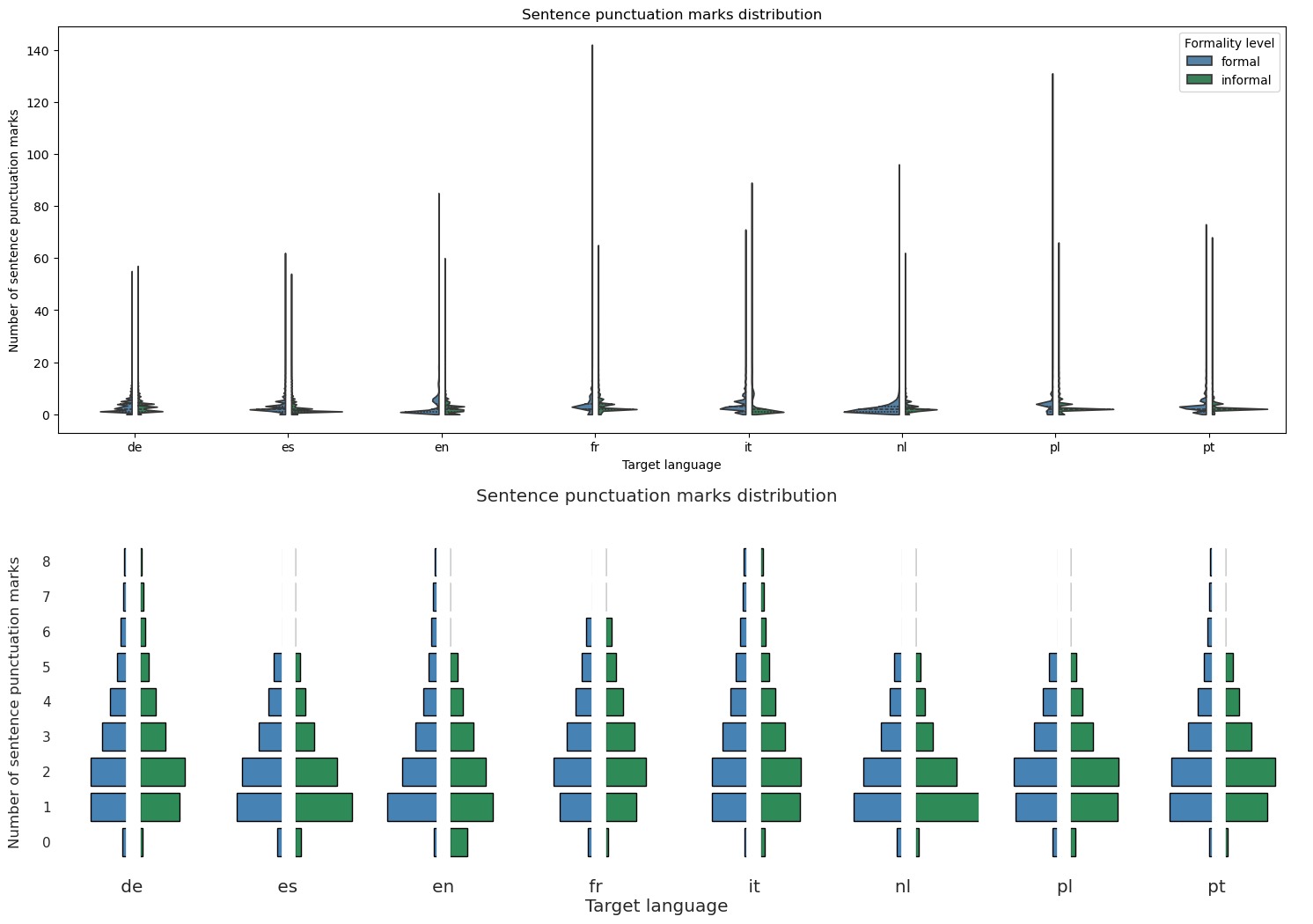}
  \caption{Plots representing the distributions of the number of punctuation signs in a sentence for a given language. The upper figure represents the distributions calculated over the original dataset. As it shows that there are some outliers, we provide the lower figure generated over a subset of texts whose lengths are between Q1 - 1.5 IQR and Q3 + 1.5 IQR (Q1=first quartile, Q3=third quartile, IQR=inter-quartile range) to focus more on the most common scenarios.}
  \label{fig:interpunction}
\end{figure*}

\begin{figure*}[ht]
  \includegraphics[width=\textwidth]{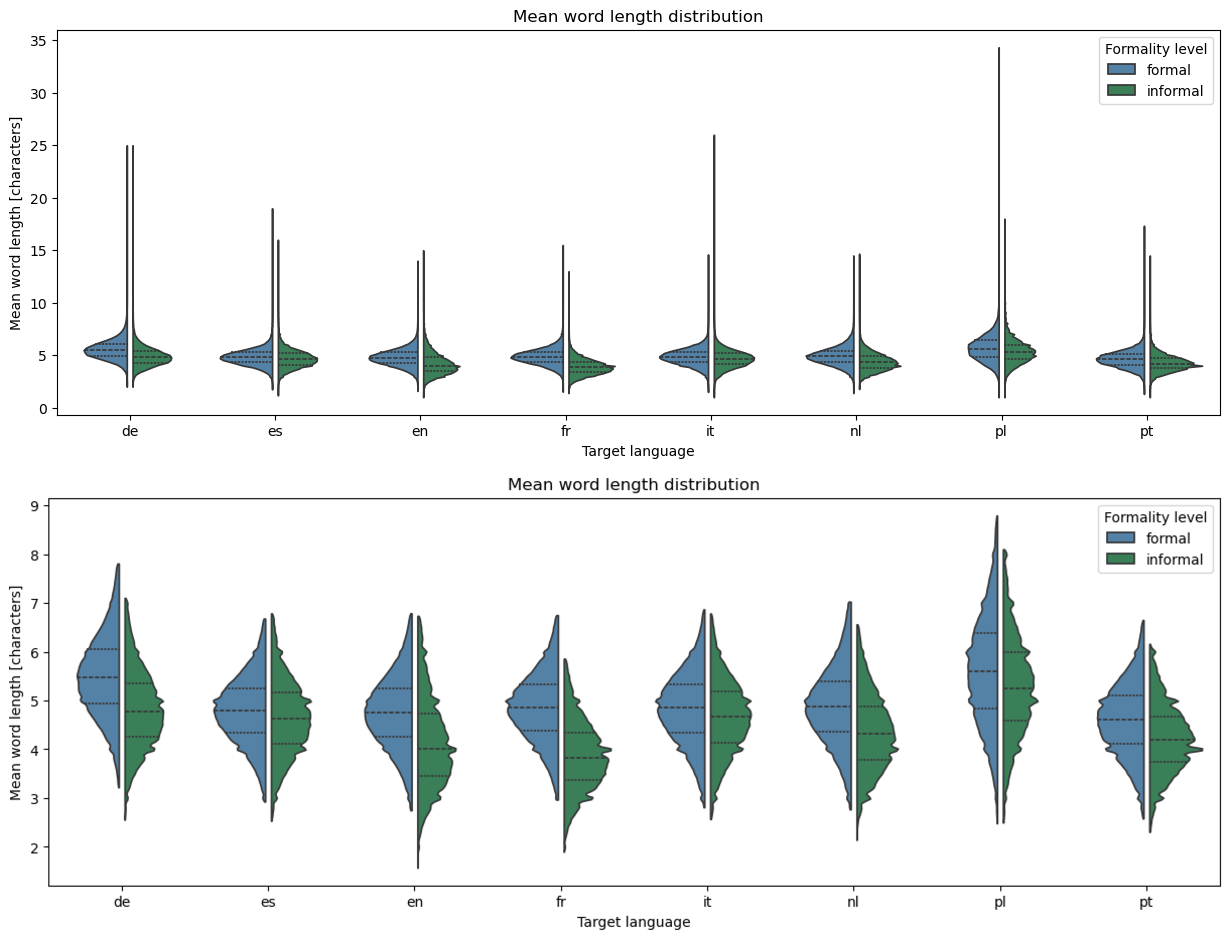}
  \caption{Plots representing the distributions of the mean word length in a given sentence per given language. The upper figure represents the distributions calculated over the original dataset. As it shows that there are some outliers, we provide the lower figure generated over a subset of texts whose lengths are between Q1 - 1.5 IQR and Q3 + 1.5 IQR (Q1=first quartile, Q3=third quartile, IQR=inter-quartile range) to focus more on the most common scenarios.}
  \label{fig:word_length}
\end{figure*}

Finally, Figures~\ref{fig:interpunction} and~\ref{fig:word_length} provide information about the punctuation distribution and word length distribution for formal and informal texts. While formal documents tend to use longer words, the distribution of punctuation marks is quite similar for both categories.

\begin{table*}
\scriptsize
    \begin{center}
\begin{tabular}{|l||*{8}{c|}}\hline
\backslashbox{Source language}{Target language}
&\makebox[4em]{German (DE)}&\makebox[4em]{English (EN)}&\makebox[4em]{Spanish (ES)}
&\makebox[4em]{French (FR)}&\makebox[3em]{Italian (IT)}&\makebox[4em]{Dutch (NL)}&\makebox[4em]{Polish (PL)}&\makebox[6em]{Portuguese (PT)}\\\hline\hline
Czech (CS) & \checkmark & \checkmark & \checkmark & \checkmark & \checkmark & \checkmark & \checkmark &  \checkmark \\\hline
Danish (DA) & \checkmark & \checkmark & \checkmark & \checkmark & \checkmark & \checkmark & \checkmark &  \checkmark \\\hline
German (DE) & NOT & \checkmark & \checkmark & \checkmark & \checkmark & \checkmark & \checkmark &  \checkmark \\\hline
English (EN) & \checkmark & NOT & \checkmark & \checkmark & \checkmark & \checkmark & \checkmark &  \checkmark \\\hline
Spanish (ES) & \checkmark & \checkmark & NOT & \checkmark & \checkmark & \checkmark & \checkmark &  \checkmark \\\hline
French (FR) & \checkmark & \checkmark & \checkmark & NOT & \checkmark & \checkmark & \checkmark &  \checkmark \\\hline
Italian (IT) & \checkmark & \checkmark & \checkmark & \checkmark & NOT & \checkmark & \checkmark &  \checkmark \\\hline
Dutch (NL) & \checkmark & \checkmark & \checkmark & \checkmark & \checkmark & NOT & \checkmark &  \checkmark \\\hline
Norwegian (NO+NB) & \checkmark & \checkmark & \checkmark & \checkmark & \checkmark & \checkmark & \checkmark &  \checkmark \\\hline
Polish (PL) & \checkmark & \checkmark & \checkmark & \checkmark & \checkmark & \checkmark & NOT &  \checkmark \\\hline
Portuguese (PT) & \checkmark & \checkmark & \checkmark & \checkmark & \checkmark & \checkmark & \checkmark &  NOT \\\hline
Russian (RU) & \checkmark & \checkmark & \checkmark & \checkmark & \checkmark & \checkmark & \checkmark &  \checkmark \\\hline
Slovak (SK) & \checkmark & \checkmark & \checkmark & \checkmark & \checkmark & \checkmark & \checkmark &  \checkmark \\\hline
Swedish (SV) & \checkmark & \checkmark & \checkmark & \checkmark & \checkmark & \checkmark & \checkmark &  \checkmark \\\hline
Ukrainian (UK) & \checkmark & \checkmark & \checkmark & \checkmark & \checkmark & \checkmark & \checkmark &  \checkmark \\\hline
\end{tabular}
\caption{Language pairs in FAME-MT. The total size of FAME-MT = 11.2 million examples (14 source languages $\cdot$ 8 target languages $\cdot$ 100,000 examples). Each cell with \checkmark represents a language pair with 100,000 examples (50\% formal translations and 50\% informal). Each cell with \texttt{NOT} represents a language pair that is not supported because source language equals target language.}
         \label{tab:mt_fame}
\end{center}
\end{table*}

\begin{table*}
\scriptsize
    \begin{center}
\begin{tabular}{|l||*{8}{c|}}\hline
\backslashbox{Source language}{Target language}
&\makebox[3em]{German (DE)}&\makebox[3em]{English (EN)}&\makebox[3em]{Spanish (ES)}
&\makebox[3em]{French (FR)}&\makebox[3em]{Italian (IT)}&\makebox[3em]{Dutch (NL)}&\makebox[6em]{Portuguese (PT)}\\\hline\hline
Czech (CS) & 53.88/\textbf{77.84} & 63.55/\textbf{83.98} & 75.05/\textbf{88.91} & 82.41/\textbf{103.41} & 78.95/\textbf{88.3} & 60.71/\textbf{84.91} & 75.93/\textbf{92.6} \\\hline
Danish (DA) & 50.76/\textbf{72.02} & 64.37/\textbf{86.56} & 65.67/\textbf{78.72} & 76.9/\textbf{100.84} & 66.95/\textbf{80.14} & 50.15/\textbf{73.15} & 62.53/\textbf{89.26} \\ \hline
German (DE) & -/- & 61.56/\textbf{83.35} & 66.11/\textbf{76.03} & 76.54/\textbf{95.52} & 65.74/\textbf{75.74} & 56.96/\textbf{72.34} & 61.56/\textbf{85.69} \\\hline
English (EN) & 49.81/\textbf{63.57} & -/- & 63.49/\textbf{72.15} & 76.55/\textbf{95.33} & 67.05/\textbf{75.14} & 57.84/\textbf{72.29} & 63.55/\textbf{83.46} \\\hline
Spanish (ES) & 49.85/\textbf{69.43} & 59.98/\textbf{92.34} & -/- & 76.91/\textbf{97.05} & 72.54/\textbf{82.05} & 57.41/\textbf{75.33} & 63.89/\textbf{84.97} \\\hline
French (FR) & 50.4/\textbf{68.33} & 58.58/\textbf{93.71} & 66.47/\textbf{78.62} & -/- & 68.77/\textbf{78.45} & 57.6/\textbf{74.74} & 66.96/\textbf{87.36} \\\hline
Italian (IT) & 49.9/\textbf{69.84} & 61.37/\textbf{91.32} & 66.9/\textbf{78.4} & 76.55/\textbf{97.37} & -/- & 56.66/\textbf{74.41} & 64.4/\textbf{86.08} \\\hline
Dutch (NL) & 51.31/\textbf{71.75} & 67.11/\textbf{94.6} & 69.67/\textbf{81.12} & 77.03/\textbf{101.03} & 70.41/\textbf{82.1} & -/- & 68.43/\textbf{90.56} \\\hline
Norwegian (NO+NB) & 51.88/\textbf{72.08} & 68.05/\textbf{89.17} & 69.1/\textbf{75.82} & 75.55/\textbf{100.69} & 74.42/\textbf{79.05} & 54.4/\textbf{62.36} &  66.94/\textbf{88.24} \\\hline
Polish (PL) & 53.12/\textbf{78.02} & 65.36/\textbf{93.55} & 71.68/\textbf{84.36} & 80.53/\textbf{103.4} & 76.89/\textbf{87.86} & 60.76/\textbf{83.15} & 74.61/\textbf{92.08} \\\hline
Portuguese (PT) & 52.12/\textbf{74.13} & 62.01/\textbf{95.72} & 70.07/\textbf{83.84} & 79.54/\textbf{101.12} & 72.92/\textbf{84.36} & 60.29/\textbf{83.69} & -/- \\\hline
Russian (RU) & 54.71/\textbf{78.11} & 60.37/\textbf{94.95} & 69.43/\textbf{86.05} & 82.92/\textbf{102.64} & 79.77/\textbf{88.9} & 62.07/\textbf{86.24} &  72.61/\textbf{94.38} \\\hline
Slovak (SK) & 52.15/\textbf{73.61} & 62.26/\textbf{90.41} & 66.31/\textbf{81.35} & 77.27/\textbf{99.28} & 65.89/\textbf{78.68} & 56.78/\textbf{76.96} & 63.21/\textbf{88} \\\hline
Swedish (SV) & 52.4/\textbf{73.12} & 65.26/\textbf{91.95} & 69.56/\textbf{80.21} & 77.06/\textbf{100.96} & 68.14/\textbf{79.95} & 57.67/\textbf{74.65} &  67.58/\textbf{90.9} \\\hline
Ukrainian (UK) & 55.63/\textbf{74.16} & 64.76/\textbf{90.88} & 70.1/\textbf{80.79} & 80.79/\textbf{96.98} & 74.69/\textbf{78.82} & 57.27/\textbf{74.29} &  64.38/\textbf{84.51} \\\hline
\textbf{Average} & 51.99/\textbf{72.57} & 63.19/\textbf{90.89} & 68.93/\textbf{80.46} & 78.33/\textbf{99.69} & 71.65/\textbf{81.4} & 57.61/\textbf{76.32} & 66.9/\textbf{88.44} \\ \hline

\end{tabular}
\caption{Flesch readability scores between each pair of languages (higher value $\rightarrow$  text easier to read). Each cell contains two numbers separated by a slash sign. The first number represents the Flesch score calculated for a random sample of 5000 formal sentences expressed in a given language pair. The second number represents an analogous score for informal target sentences. In each sample, for each language pair, informal texts are easier to read (higher scores marked in bold).}
         \label{tab:flesch}
\end{center}
\end{table*}

\begin{table*}
\scriptsize
    \begin{center}

\begin{tabular}{|r||c|c|c||c|c|c|} \hline
\multirow{2}{*}{} & \multicolumn{3}{c||}{\textbf{Formal target}} & \multicolumn{3}{c|}{\textbf{Informal target}} \\
\textbf{Source and target languages} & \textbf{Formal source} & \textbf{Neutral source} & \textbf{Informal source} & \textbf{Informal source} &  \textbf{Neutral source} & \textbf{Formal source}  \\ \hline

English (EN) $\rightarrow$ German (DE) & 35070 \textbf{(70.14\%)} & 12390 \textbf{(24.78\%)} & 2540 \textbf{(5.08\%)} & 5281 \textbf{(10.56\%)} & 14562 \textbf{(29.12\%)}& 30157 \textbf{(60.31\%)} \\ \hline
Spanish (ES) $\rightarrow$ German (DE) & 17555 \textbf{(35.11\%)} & 21012 \textbf{(42.02\%)} & 11433 \textbf{(22.87\%)} & 32253 \textbf{(64.51\%)} & 11288 \textbf{(22.58\%)}& 6459 \textbf{(12.92\%)} \\ \hline
French (FR) $\rightarrow$ German (DE) & 33929 \textbf{(67.86\%)} & 15242 \textbf{(30.48\%)} & 829 \textbf{(1.66\%)} & 18925 \textbf{(37.85\%)} & 9963 \textbf{(19.93\%)}& 21112 \textbf{(42.22\%)} \\ \hline
Italian (IT) $\rightarrow$ German (DE) & 5496 \textbf{(10.99\%)} & 31903 \textbf{(63.81\%)} & 12601 \textbf{(25.2\%)} & 30364 \textbf{(60.73\%)} & 16867 \textbf{(33.73\%)}& 2769 \textbf{(5.54\%)} \\ \hline
Polish (PL) $\rightarrow$ German (DE) & 6498 \textbf{(13.0\%)} & 25771 \textbf{(51.54\%)} & 17731 \textbf{(35.46\%)} & 35236 \textbf{(70.47\%)} & 12164 \textbf{(24.33\%)}& 2600 \textbf{(5.2\%)} \\ \hline
Portugal (PT) $\rightarrow$ German (DE) & 29499 \textbf{(59.0\%)} & 19213 \textbf{(38.43\%)} & 1288 \textbf{(2.58\%)} & 12310 \textbf{(24.62\%)} & 12770 \textbf{(25.54\%)}& 24920 \textbf{(49.84\%)} \\ \hline
Dutch (NL) $\rightarrow$ German (DE) & 21803 \textbf{(43.61\%)} & 15463 \textbf{(30.93\%)} & 12734 \textbf{(25.47\%)} & 35478 \textbf{(70.96\%)} & 10576 \textbf{(21.15\%)}& 3946 \textbf{(7.89\%)} \\ \hline
German (DE) $\rightarrow$ English (EN) & 8444 \textbf{(16.89\%)} & 39829 \textbf{(79.66\%)} & 1727 \textbf{(3.45\%)} & 2783 \textbf{(5.57\%)} & 42584 \textbf{(85.17\%)}& 4633 \textbf{(9.27\%)} \\ \hline
Spanish (ES) $\rightarrow$ English (EN) & 8489 \textbf{(16.98\%)} & 36997 \textbf{(73.99\%)} & 4514 \textbf{(9.03\%)} & 9080 \textbf{(18.16\%)} & 35870 \textbf{(71.74\%)}& 5050 \textbf{(10.1\%)} \\ \hline
French (FR) $\rightarrow$ English (EN) & 8760 \textbf{(17.52\%)} & 40154 \textbf{(80.31\%)} & 1086 \textbf{(2.17\%)} & 5765 \textbf{(11.53\%)} & 36896 \textbf{(73.79\%)}& 7339 \textbf{(14.68\%)} \\ \hline
Italian (IT) $\rightarrow$ English (EN) & 4204 \textbf{(8.41\%)} & 40544 \textbf{(81.09\%)} & 5252 \textbf{(10.5\%)} & 7748 \textbf{(15.5\%)} & 39355 \textbf{(78.71\%)}& 2897 \textbf{(5.79\%)} \\ \hline
Polish (PL) $\rightarrow$ English (EN) & 2821 \textbf{(5.64\%)} & 38964 \textbf{(77.93\%)} & 8215 \textbf{(16.43\%)} & 14810 \textbf{(29.62\%)} & 32751 \textbf{(65.5\%)}& 2439 \textbf{(4.88\%)} \\ \hline
Portugal (PT) $\rightarrow$ English (EN) & 13104 \textbf{(26.21\%)} & 35920 \textbf{(71.84\%)} & 976 \textbf{(1.95\%)} & 3234 \textbf{(6.47\%)} & 35328 \textbf{(70.66\%)}& 11438 \textbf{(22.88\%)} \\ \hline
Dutch (NL) $\rightarrow$ English (EN) & 5585 \textbf{(11.17\%)} & 37794 \textbf{(75.59\%)} & 6621 \textbf{(13.24\%)} & 8831 \textbf{(17.66\%)} & 39525 \textbf{(79.05\%)}& 1644 \textbf{(3.29\%)} \\ \hline
German (DE) $\rightarrow$ Spanish (ES) & 21142 \textbf{(42.28\%)} & 26750 \textbf{(53.5\%)} & 2108 \textbf{(4.22\%)} & 17301 \textbf{(34.6\%)} & 10018 \textbf{(20.04\%)}& 22681 \textbf{(45.36\%)} \\ \hline
English (EN) $\rightarrow$ Spanish (ES) & 36314 \textbf{(72.63\%)} & 10132 \textbf{(20.26\%)} & 3554 \textbf{(7.11\%)} & 8148 \textbf{(16.3\%)} & 14101 \textbf{(28.2\%)}& 27751 \textbf{(55.5\%)} \\ \hline
French (FR) $\rightarrow$ Spanish (ES) & 16087 \textbf{(32.17\%)} & 32477 \textbf{(64.95\%)} & 1436 \textbf{(2.87\%)} & 13417 \textbf{(26.83\%)} & 10920 \textbf{(21.84\%)}& 25663 \textbf{(51.33\%)} \\ \hline
Italian (IT) $\rightarrow$ Spanish (ES) & 15430 \textbf{(30.86\%)} & 27140 \textbf{(54.28\%)} & 7430 \textbf{(14.86\%)} & 27970 \textbf{(55.94\%)} & 18995 \textbf{(37.99\%)}& 3035 \textbf{(6.07\%)} \\ \hline
Polish (PL) $\rightarrow$ Spanish (ES) & 7099 \textbf{(14.2\%)} & 30481 \textbf{(60.96\%)} & 12420 \textbf{(24.84\%)} & 34870 \textbf{(69.74\%)} & 12181 \textbf{(24.36\%)}& 2949 \textbf{(5.9\%)} \\ \hline
Portugal (PT) $\rightarrow$ Spanish (ES) & 34107 \textbf{(68.21\%)} & 14891 \textbf{(29.78\%)} & 1002 \textbf{(2.0\%)} & 9874 \textbf{(19.75\%)} & 12240 \textbf{(24.48\%)}& 27886 \textbf{(55.77\%)} \\ \hline
Dutch (NL) $\rightarrow$ Spanish (ES) & 14564 \textbf{(29.13\%)} & 27806 \textbf{(55.61\%)} & 7630 \textbf{(15.26\%)} & 31427 \textbf{(62.85\%)} & 9942 \textbf{(19.88\%)}& 8631 \textbf{(17.26\%)} \\ \hline
German (DE) $\rightarrow$ French (FR) & 36371 \textbf{(72.74\%)} & 7966 \textbf{(15.93\%)} & 5663 \textbf{(11.33\%)} & 29805 \textbf{(59.61\%)} & 14752 \textbf{(29.5\%)}& 5443 \textbf{(10.89\%)} \\ \hline
English (EN) $\rightarrow$ French (FR) & 33675 \textbf{(67.35\%)} & 12696 \textbf{(25.39\%)} & 3629 \textbf{(7.26\%)} & 15148 \textbf{(30.3\%)} & 12471 \textbf{(24.94\%)}& 22381 \textbf{(44.76\%)} \\ \hline
Spanish (ES) $\rightarrow$ French (FR) & 17940 \textbf{(35.88\%)} & 15202 \textbf{(30.4\%)} & 16858 \textbf{(33.72\%)} & 30891 \textbf{(61.78\%)} & 13582 \textbf{(27.16\%)}& 5527 \textbf{(11.05\%)} \\ \hline
Italian (IT) $\rightarrow$ French (FR) & 5948 \textbf{(11.9\%)} & 27005 \textbf{(54.01\%)} & 17047 \textbf{(34.09\%)} & 29517 \textbf{(59.03\%)} & 16683 \textbf{(33.37\%)}& 3800 \textbf{(7.6\%)} \\ \hline
Polish (PL) $\rightarrow$ French (FR) & 8272 \textbf{(16.54\%)} & 18179 \textbf{(36.36\%)} & 23549 \textbf{(47.1\%)} & 36702 \textbf{(73.4\%)} & 10877 \textbf{(21.75\%)}& 2421 \textbf{(4.84\%)} \\ \hline
Portugal (PT) $\rightarrow$ French (FR) & 33897 \textbf{(67.79\%)} & 13684 \textbf{(27.37\%)} & 2419 \textbf{(4.84\%)} & 14252 \textbf{(28.5\%)} & 14079 \textbf{(28.16\%)}& 21669 \textbf{(43.34\%)} \\ \hline
Dutch (NL) $\rightarrow$ French (FR) & 23659 \textbf{(47.32\%)} & 7537 \textbf{(15.07\%)} & 18804 \textbf{(37.61\%)} & 34132 \textbf{(68.26\%)} & 14393 \textbf{(28.79\%)}& 1475 \textbf{(2.95\%)} \\ \hline
German (DE) $\rightarrow$ Italian (IT) & 17323 \textbf{(34.65\%)} & 31410 \textbf{(62.82\%)} & 1267 \textbf{(2.5\%)} & 15333 \textbf{(30.67\%)} & 8605 \textbf{(17.21\%)}& 26062 \textbf{(52.12\%)} \\ \hline
English (EN) $\rightarrow$ Italian (IT) & 34491 \textbf{(68.98\%)} & 10309 \textbf{(20.62\%)} & 5200 \textbf{(10.4\%)} & 8882 \textbf{(17.76\%)} & 12870 \textbf{(25.74\%)}& 28248 \textbf{(56.5\%)} \\ \hline
Spanish (ES) $\rightarrow$ Italian (IT) & 28522 \textbf{(57.04\%)} & 16438 \textbf{(32.88\%)} & 5040 \textbf{(10.08\%)} & 30705 \textbf{(61.41\%)} & 9555 \textbf{(19.11\%)}& 9740 \textbf{(19.48\%)} \\ \hline
French (FR) $\rightarrow$ Italian (IT) & 11666 \textbf{(23.33\%)} & 36215 \textbf{(72.43\%)} & 2119 \textbf{(4.24\%)} & 13321 \textbf{(26.64\%)} & 10083 \textbf{(20.17\%)}& 26596 \textbf{(53.19\%)} \\ \hline
Polish (PL) $\rightarrow$ Italian (IT) & 9932 \textbf{(19.86\%)} & 27943 \textbf{(55.89\%)} & 12125 \textbf{(24.25\%)} & 34979 \textbf{(69.96\%)} & 12180 \textbf{(24.36\%)}& 2841 \textbf{(5.68\%)} \\ \hline
Portugal (PT) $\rightarrow$ Italian (IT) & 31358 \textbf{(62.72\%)} & 17083 \textbf{(34.17\%)} & 1559 \textbf{(3.12\%)} & 10006 \textbf{(20.01\%)} & 11631 \textbf{(23.26\%)}& 28363 \textbf{(56.73\%)} \\ \hline
Dutch (NL) $\rightarrow$ Italian (IT) & 8003 \textbf{(16.01\%)} & 34786 \textbf{(69.57\%)} & 7211 \textbf{(14.42\%)} & 29321 \textbf{(58.64\%)} & 9900 \textbf{(19.8\%)}& 10779 \textbf{(21.56\%)} \\ \hline
German (DE) $\rightarrow$ Polish (PL) & 20330 \textbf{(40.66\%)} & 26744 \textbf{(53.49\%)} & 2926 \textbf{(5.85\%)} & 12171 \textbf{(24.34\%)} & 20941 \textbf{(41.88\%)}& 16888 \textbf{(33.78\%)} \\ \hline
English (EN) $\rightarrow$ Polish (PL) & 31044 \textbf{(62.09\%)} & 10617 \textbf{(21.23\%)} & 8339 \textbf{(16.68\%)} & 13598 \textbf{(27.2\%)} & 11868 \textbf{(23.74\%)}& 24534 \textbf{(49.07\%)} \\ \hline
Spanish (ES) $\rightarrow$ Polish (PL) & 16357 \textbf{(32.71\%)} & 28082 \textbf{(56.16\%)} & 5561 \textbf{(11.12\%)} & 18445 \textbf{(36.89\%)} & 18661 \textbf{(37.32\%)}& 12894 \textbf{(25.79\%)} \\ \hline
French (FR) $\rightarrow$ Polish (PL) & 19355 \textbf{(38.71\%)} & 27293 \textbf{(54.59\%)} & 3352 \textbf{(6.7\%)} & 13362 \textbf{(26.72\%)} & 21551 \textbf{(43.1\%)}& 15087 \textbf{(30.17\%)} \\ \hline
Italian (IT) $\rightarrow$ Polish (PL) & 11410 \textbf{(22.82\%)} & 32094 \textbf{(64.19\%)} & 6496 \textbf{(12.99\%)} & 20093 \textbf{(40.19\%)} & 26253 \textbf{(52.51\%)}& 3654 \textbf{(7.31\%)} \\ \hline
Portugal (PT) $\rightarrow$ Polish (PL) & 20408 \textbf{(40.82\%)} & 27196 \textbf{(54.39\%)} & 2396 \textbf{(4.79\%)} & 8348 \textbf{(16.7\%)} & 22168 \textbf{(44.34\%)}& 19484 \textbf{(38.97\%)} \\ \hline
Dutch (NL) $\rightarrow$ Polish (PL) & 11248 \textbf{(22.5\%)} & 28816 \textbf{(57.63\%)} & 9936 \textbf{(19.87\%)} & 22401 \textbf{(44.8\%)} & 22483 \textbf{(44.97\%)}& 5116 \textbf{(10.23\%)} \\ \hline
German (DE) $\rightarrow$ Portugal (PT) & 22322 \textbf{(44.64\%)} & 20169 \textbf{(40.34\%)} & 7509 \textbf{(15.02\%)} & 27060 \textbf{(54.12\%)} & 15757 \textbf{(31.51\%)}& 7183 \textbf{(14.37\%)} \\ \hline
English (EN) $\rightarrow$ Portugal (PT) & 33023 \textbf{(66.05\%)} & 11556 \textbf{(23.11\%)} & 5421 \textbf{(10.84\%)} & 14020 \textbf{(28.04\%)} & 12330 \textbf{(24.66\%)}& 23650 \textbf{(47.3\%)} \\ \hline
Spanish (ES) $\rightarrow$ Portugal (PT) & 21487 \textbf{(42.97\%)} & 14667 \textbf{(29.33\%)} & 13846 \textbf{(27.69\%)} & 31533 \textbf{(63.07\%)} & 14485 \textbf{(28.97\%)}& 3982 \textbf{(7.96\%)} \\ \hline
French (FR) $\rightarrow$ Portugal (PT) & 21263 \textbf{(42.53\%)} & 22366 \textbf{(44.73\%)} & 6371 \textbf{(12.74\%)} & 25245 \textbf{(50.49\%)} & 15808 \textbf{(31.62\%)}& 8947 \textbf{(17.89\%)} \\ \hline
Italian (IT) $\rightarrow$ Portugal (PT) & 10367 \textbf{(20.73\%)} & 24452 \textbf{(48.9\%)} & 15181 \textbf{(30.36\%)} & 29337 \textbf{(58.67\%)} & 17779 \textbf{(35.56\%)}& 2884 \textbf{(5.77\%)} \\ \hline
Polish (PL) $\rightarrow$ Portugal (PT) & 5696 \textbf{(11.39\%)} & 20486 \textbf{(40.97\%)} & 23818 \textbf{(47.64\%)} & 35852 \textbf{(71.7\%)} & 11792 \textbf{(23.58\%)}& 2356 \textbf{(4.71\%)} \\ \hline
Dutch (NL) $\rightarrow$ Portugal (PT) & 10915 \textbf{(21.83\%)} & 19712 \textbf{(39.42\%)} & 19373 \textbf{(38.75\%)} & 33186 \textbf{(66.37\%)} & 15465 \textbf{(30.93\%)}& 1349 \textbf{(2.7\%)} \\ \hline
German (DE) $\rightarrow$ Dutch (NL) & 40742 \textbf{(81.48\%)} & 7124 \textbf{(14.25\%)} & 2134 \textbf{(4.27\%)} & 16806 \textbf{(33.61\%)} & 11831 \textbf{(23.66\%)}& 21363 \textbf{(42.73\%)} \\ \hline
English (EN) $\rightarrow$ Dutch (NL) & 36558 \textbf{(73.12\%)} & 10737 \textbf{(21.47\%)} & 2705 \textbf{(5.41\%)} & 8698 \textbf{(17.4\%)} & 12496 \textbf{(24.99\%)}& 28806 \textbf{(57.61\%)} \\ \hline
Spanish (ES) $\rightarrow$ Dutch (NL) & 22760 \textbf{(45.52\%)} & 15556 \textbf{(31.11\%)} & 11684 \textbf{(23.37\%)} & 28258 \textbf{(56.52\%)} & 13826 \textbf{(27.65\%)}& 7916 \textbf{(15.83\%)} \\ \hline
French (FR) $\rightarrow$ Dutch (NL) & 41421 \textbf{(82.84\%)} & 7746 \textbf{(15.49\%)} & 833 \textbf{(1.67\%)} & 14766 \textbf{(29.53\%)} & 9928 \textbf{(19.86\%)}& 25306 \textbf{(50.61\%)} \\ \hline
Italian (IT) $\rightarrow$ Dutch (NL) & 6583 \textbf{(13.17\%)} & 28871 \textbf{(57.74\%)} & 14546 \textbf{(29.09\%)} & 25372 \textbf{(50.74\%)} & 20634 \textbf{(41.27\%)}& 3994 \textbf{(7.99\%)} \\ \hline
Polish (PL) $\rightarrow$ Dutch (NL) & 10729 \textbf{(21.46\%)} & 21618 \textbf{(43.24\%)} & 17653 \textbf{(35.31\%)} & 33024 \textbf{(66.05\%)} & 13191 \textbf{(26.38\%)}& 3785 \textbf{(7.57\%)} \\ \hline
Portugal (PT) $\rightarrow$ Dutch (NL) & 35708 \textbf{(71.42\%)} & 13140 \textbf{(26.28\%)} & 1152 \textbf{(2.3\%)} & 11072 \textbf{(22.14\%)} & 13441 \textbf{(26.88\%)}& 25487 \textbf{(50.97\%)} \\ \hline

\end{tabular}
\caption{Relations between source sentence formality and target sentence formality in \textbf{FAME-MT} determined using our classifiers. For each formal and informal target sentence, the classifier is used to determine the formality of the corresponding source sentence. Then, the number of source sentences classified as formal, informal, and neutral is reported for each target category.}
         \label{tab:confusion_matrix}
\end{center}
\end{table*}

\end{document}